\pdfoutput=1
\documentclass[pdftex,table,xcdraw,11pt]{article}

\usepackage[]{ACL2023}

\usepackage{times}

\usepackage{longtable}
\usepackage{latexsym}
\usepackage{enumitem}
\usepackage{tcolorbox}
\usepackage{booktabs}
\usepackage{xcolor}
\usepackage{amsmath}
\usepackage{algorithm}
\usepackage{multirow}
\usepackage{algpseudocode}
\usepackage{setspace}
\usepackage{pifont}
\usepackage{framed}
\usepackage{amsmath}
\usepackage{rotating} 
\usepackage{colortbl}
\usepackage{multicol}
\usepackage{pgfplots}
\usepackage{colortbl}
\usepackage{float} 
\usepackage[utf8]{inputenc}
\usepackage{array}
\pgfplotsset{compat=1.17}
\usepgfplotslibrary{colormaps}

\usepackage[normalem]{ulem}
\useunder{\uline}{\ul}{}
\usepackage[T1]{fontenc}

\usepackage{kotex}
\let\oldul\ul
\let\ul\relax
\usepackage{soul}
\let\ul\oldul

\usepackage{inconsolata}
\usepackage{subcaption}
\usepackage{graphicx}
\tcbuselibrary{breakable}
\usepackage{hyperref}
\definecolor{mygreen}{RGB}{0, 128, 0}

\newcommand{\hllow}[1]{\sethlcolor{red!10}\hl{#1}}

\newcommand{\Interviewer}{\textcolor{green!50!black}{\texttt{Interviewer}}}

\newcommand{\Interviewee}{\textcolor{orange!100!red}{\texttt{Interviewee}}}

\title{\raisebox{-0.02cm}
{\textsc{\LARGE LLM-as-an-Interviewer}}:\\ Beyond Static Testing Through Dynamic LLM Evaluation}

\author{Eunsu Kim$^{1}$ \quad Juyoung Suk$^{1}$\quad Seungone Kim$^{2}$ \quad Niklas Muennighoff$^{3,4}$ \\ \quad \textbf{Dongkwan Kim}$^{1}$ \quad \textbf{Alice Oh}$^{1}$
  \\\\
$^1$KAIST, $^2$Carnegie Mellon University, $^3$Stanford University, $^4$Contextual AI
\\\texttt{\href{mailto:kes0317@kaist.ac.kr}{\color{black}{kes0317@kaist.ac.kr}}, \href{mailto:alice.oh@kaist.edu}{\color{black}{alice.oh@kaist.edu}}}}

\usepackage{amsmath,amsfonts,bm}

\def\eqref#1{equation~\ref{#1}}

\def\1{\bm{1}}

\DeclareMathAlphabet{\mathsfit}{\encodingdefault}{\sfdefault}{m}{sl}
\SetMathAlphabet{\mathsfit}{bold}{\encodingdefault}{\sfdefault}{bx}{n}

\begin{document}
\maketitle

\begin{abstract}
We introduce \textbf{\textsc{LLM-as-an-Interviewer}}, a novel paradigm for evaluating large language models (LLMs). This approach leverages multi-turn interactions where the LLM interviewer actively provides feedback on responses and poses follow-up questions to the evaluated LLM. At the start of the interview, the LLM interviewer dynamically modifies datasets to generate initial questions, mitigating data contamination.
We apply the \textbf{\text{LLM-as-an-Interviewer}} framework to evaluate six models on the reasoning, factuality and instruction-following tasks. Our results show that the framework effectively provides insights into LLM performance, including the quality of initial responses, adaptability to feedback, and ability to address follow-up queries like clarification or additional knowledge requests. The framework also addresses key limitations of conventional methods like LLM-as-a-Judge, including verbosity bias and inconsistency across runs.
Finally, we propose the \textbf{Interview Report}, which aggregates insights from the interview process, providing examples and a comprehensive analysis of the LLM’s strengths and weaknesses. This report offers a detailed snapshot of the model's real-world applicability\footnote{The code for our framework is publicly available at \href{https://github.com/interview-eval}{https://github.com/interview-eval}}\footnote{We provide a general Python library applicable to various tasks: \href{https://pypi.org/project/interview-eval/}{https://pypi.org/project/interview-eval/}}.
\end{abstract}
\section{Introduction}

With large language models (LLMs) becoming increasingly proficient in generating fluent free-form responses, it has become crucial to properly assess their capabilities and limitations~\citep{liang2022holistic,chang2024survey}. Recently, LLM-as-a-judge has emerged as a promising framework for automatic free-form response evaluation. Compared to traditional lexical matching-based metrics (\textit{e.g.}, ROUGE~\citep{lin2004rouge}, BLEU~\citep{papineni2002bleu}) or embedding-based metrics (\textit{e.g.}, BERTScore~\citep{zhang2019bertscore}), previous works on LLM-as-a-Judge have reported higher correlations with human judgments~\citep{chiang2023can,zheng2023judging,dubois2024length}.

Despite its potential, the LLM-as-a-Judge framework faces several practical limitations that hinder its widespread adoption, primarily due to its \textit{static nature}~\citep{li2024generation,gu2024survey}. First, using a \textit{fixed set of test inputs} raises concerns about data contamination~\citep{sainz2023nlp,zhou2023don,oren2023proving}, where the evaluated models may achieve high scores on instances encountered during training. Second, \textit{single-turn interactions} fail to thoroughly probe a model's true comprehension~\citep{li2019acute,wang2023mint,kwan2024mt}. For instance, the judge model may assess the confined performance of LLMs distant from the use case, be influenced by superficial factors (e.g., favoring longer responses), and exhibit high variance across runs.

\begin{figure*}[ht]
    \centering
    \includegraphics[width=\textwidth]{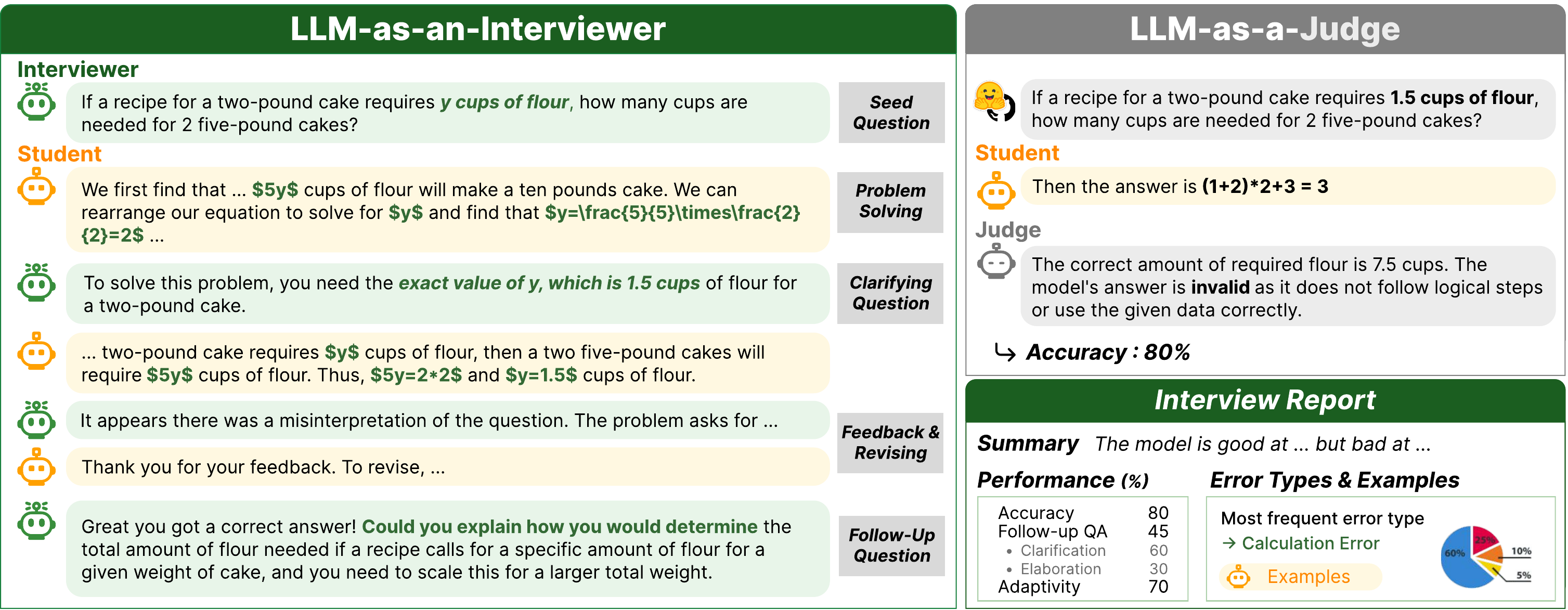}
    \caption{\textbf{Overview of \text{LLM-as-an-Interviewer}}. \textit{LLM-as-an-Interviewer} aims to assess LLMs by 1) constructing seed questions of interviews based on existing benchmark datasets, and 2) providing feedback or asking additional follow-up questions. Unlike evaluation being limited to a single score with LLM-as-a-Judge, the \textit{Interview Report} provided by our framework offers a snapshot of what the model excels at and where it falls short, along with scores for various abilities and examples.}
    \label{fig:overview}
\end{figure*}

In this study, we propose \textbf{\textsc{LLM-as-an-Interviewer}}, a new paradigm for evaluating LLMs. Inspired by human interviews, this approach starts with general questions but dynamically adapts by posing different types of questions based on the model’s responses. As shown in Figure~\ref{fig:overview}, the LLM interviewer plays three key roles: (1) \text{Question Modification}, adapting benchmark datasets to generate diverse and challenging initial interview questions; (2) \text{Providing Feedback}, guiding the model to refine its responses; and (3) \text{Generating Follow-up Questions}, exploring related concepts through clarification requests or additional explanations. This dynamic evaluation reveals behaviors that static benchmarks cannot capture, such as the model’s ability to improve through feedback and provide more detailed explanations.


We demonstrate the efficacy of our framework through experiments on three tasks: Reasoning, Factuality, and Instruction Following. In \S\ref{sec:benchmarking}, we evaluate six different models using GPT-4o as the interviewer. We demonstrate the impact of providing feedback on the model's performance and suggest that follow-up questions can offer deeper insights into its behavior. These follow-ups can help uncover failure reasons or assess performance on additional requests. Furthermore, in \S~\ref{sec:DataContam} and \S~\ref{sec:reliable}, we show that our interactive and dynamic framework addresses key limitations of static evaluations, such as data contamination, verbosity bias in LLM evaluators, and high variance across multiple runs.

Moreover, we find that the extended interaction between the LLM interviewer  and interviewee contains rich information that reveals the LLM interviewee's strengths and limitations. Based on this, we introduce a new evaluation protocol, \textbf{Interview Report}, which summarizes the interaction into a structured format.

Our contributions are as follows:
\vspace{-1mm}
\begin{itemize}
\item Introduce \textbf{LLM-as-an-Interviewer}, a novel evaluation paradigm that mimics the dynamic nature of how humans evaluate humans.
\item Demonstrate that \textbf{LLM-as-an-Interviewer} mitigates several limitations of traditional evaluation approaches, such as verbosity bias, data contamination, and high variance.
\item Propose the \textbf{Interview Report}, offering a detailed snapshot of an LLM's capabilities, including examples and its common errors.
\end{itemize}

\section{Related Works}
\subsection{LLM-based Evaluation}
LLM-as-a-Judge is widely used for text evaluation, providing more human-aligned assessments than traditional methods like BLEU and ROUGE, while addressing scalability and consistency issues in human-based approaches~\cite{gu2024survey}. However, LLM-as-a-Judge faces reliability issues, including self-enhancement bias, sensitivity to response length and order, and low self-consistency. To enhance reliability, \citet{alpaca_eval} and \citet{dubois2024length} mitigate biases related to ordering and length by adjusting positions or win rates. Additionally, works such as \citet{pandalm, prometheus} train open-sourced critique models to tackle the inherent inconsistencies of closed LLMs, ensuring reproducibility.

LLM-based evaluation is also used to simulate multi-turn interactions between evaluators and models in certain tasks. For instance, \citet{wang2023mint} evaluate LLMs by simulating user-model interactions in contexts where users provide feedback during tool usage. Similarly, \citet{yu-etal-2024-kieval} simulate knowledge-focused dialogues in multiple interactions to mitigate data contamination issues.
\citet{li-etal-2023-interview} share a motivation similar to ours, as they are also inspired by human interviews. However, their approach is tailored to Conversational Question Answering tasks and emphasizes generating new questions during evaluation.

LLM-as-an-Interviewer is a generalized benchmark that mimics a multiple interview process to better assess a model's capabilities. Our framework enables the simulation of diverse interactions, such as giving feedback and asking follow-up questions, which align much more closely with how humans use models in real scenarios. 

\subsection{Data contamination in LLMs}

LLMs are trained on large corpora that can be contaminated with benchmark data~\citep{dodge2021documentinglargewebtextcorpora,soldaini2024dolmaopencorpustrillion}, undermining benchmark reliability~\citep{zhou2023dontmakellmevaluation}. This has sparked interest in \textbf{contamination detection}~\citep{magar2022datacontaminationmemorizationexploitation}. \citet{shi2024detectingpretrainingdatalarge} and \citet{oren2023provingtestsetcontamination} propose methods to estimate the likelihood of text in LLM pretraining data using only API access, as is common for frontier models\citep{openai2024gpt4technicalreport}. This has led to further research on contamination~\citep{yax2024assessingcontaminationlargelanguage,deng2024investigatingdatacontaminationmodern,jiang2024investigatingdatacontaminationpretraining,dekoninck2024constatperformancebasedcontaminationdetection,xu2024benchmarkingbenchmarkleakagelarge}, including methods to evade detection by training on rephrased benchmark data~\citep{dekoninck2024evadingdatacontaminationdetection}.

Overall, efforts to prevent contamination through detection have had limited success, prompting researchers to focus on \textbf{contamination mitigation} in benchmarks. This typically involves dynamic evaluation in two ways: \textbf{(1)} The \textit{evaluation data is dynamic}, such as crawling new instances live~\citep{white2024livebenchchallengingcontaminationfreellm,jain2024livecodebenchholisticcontaminationfree} or generating new instances on the fly~\citep{wang2024benchmarkselfevolvingmultiagentframework,li2024autobenchercreatingsalientnovel}; \textbf{(2)} The \textit{evaluator is dynamic}, as in AlpacaEval~\citep{alpaca_eval} and MT-Bench~\citep{zheng2023judgingllmasajudgemtbenchchatbot}, where the same samples are reused but evaluation depends on a dynamic LLM rating completions instead of ground-truth solutions. Some benchmarks, like Chatbot Arena~\citep{chiang2024chatbotarenaopenplatform}, combine dynamic data and evaluators, with users creating and rating responses on the fly.


LLM-as-an-Interviewer is a benchmark that mitigates contamination through both dynamic evaluation data and a dynamic evaluator. The LLM interviewer rephrases questions and proposes novel follow-ups while also evaluating the model’s responses and problem-solving process. This setup makes cheating via contamination difficult. Additionally, it enables fast, low-cost evaluation in minutes, unlike Chatbot Arena, which requires human access and takes days to yield statistically significant results at a higher cost.






\section{\textsc{LLM-as-an-Interviewer}}
We introduce the \textsc{LLM-as-an-Interviewer}, which simulates an interview where one LLM dynamically evaluates another. The LLM being evaluated is referred to as the \texttt{\Interviewee{}}, while the LLM conducting the evaluation is referred to as the \texttt{\Interviewer{}}. In this section, we describe (1) the overall interview framework (\S~\ref{sec:interview_process}) and (2) the Interview Report, which summarizes and presents the results of the interview (\S~\ref{sec:interview_report}).

\subsection{Interview Framework}
\label{sec:interview_process}
\begin{figure}[ht]
    \centering
    \includegraphics[width=0.9\columnwidth]{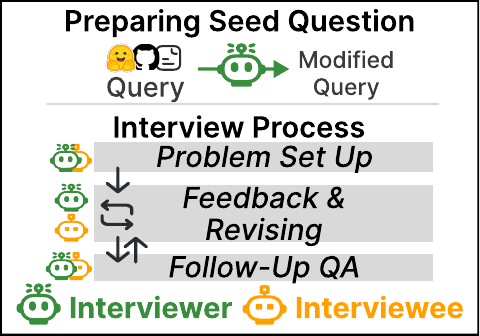}
    \caption{Workflow of LLM-as-an-Interviewer}
    \label{fig:process}
\end{figure}
We design an interview framework for LLMs as illustrated in Figure~\ref{fig:process}.
The \Interviewer{} plays three main roles throughout the interview process: (1) \textbf{Question Modification}, (2) \textbf{Providing Feedback}, and (3) \textbf{Generating Follow-up Questions}. This structure ensures a comprehensive evaluation of the \Interviewee{}’s capabilities. These roles are \textbf{modular and pluggable}, allowing users to enable or disable specific roles based on their evaluation needs. Algorithm \ref{interview_algorithm} provides the implementation of the interview process.

\subsubsection{Question Modification}
\label{sec:question_modification}
Before conducting the interview, the \Interviewer{} prepares seed questions by \textbf{modifying queries from existing benchmark datasets}. This ensures diverse scenarios that align with established evaluation standards while preventing data contamination. This approach mirrors human interviewers, who often tweak existing questions or introduce new challenges.
In our experiment, we employ two query modification strategies, prompting the model to modify queries. Each strategy is designed to enhance question diversity while maintaining ease of real-time adjustments and comparable complexity. We provide a detailed description of these methods in Appendix~\ref{appendix:experiment_1}.

Our primary goal in query modification is to mitigate data contamination and enable evaluation of multiple models using the same modified questions.
We recommend that users without contamination concerns or those preferring direct comparison with original questions may avoid query modification.

\subsubsection{Providing Feedback}
\label{sec:feedback}
During the interview process, the \Interviewer{} evaluates the response from the \Interviewee{} and provides \textbf{constructive feedback}. Providing feedback aims to guide the \Interviewee{} in identifying errors, refining the response, and improving solutions. This phase closely resembles real-world LLM interaction scenarios, where LLMs iteratively refine their responses based on user-generated feedback~\cite{wang2023mint}.

\subsubsection{{Generating Follow-up Questions}}
\label{sec:followup_questions}
After assessing the response to the initial question, the \Interviewer{} poses \textbf{follow-up questions} to further evaluate the \Interviewee{}’s ability. The purpose of follow-up questions is to gain deeper insights into the model's behavior, such as uncovering reasons for failures or assessing knowledge omitted in the initial response of the \Interviewee{} model, rather than directly comparing the accuracy across models. Follow-up questioning is crucial in many interview processes (e.g., Google coding interviews) and real-world LLM applications~\citep{bai-etal-2024-mt}.
Table~\ref{table:real_ex_feedbac_followup} provides follow-up type examples for the Reasoning, Knowledge, and Instruction Following tasks.

\subsection{Interview Report} 
\label{sec:interview_report}
The Interview Report, generated as a result of the interview process, includes (1) Performance Scores, (2) Error Analysis and Examples, and (3) Comprehensive Summary. The real examples of Interview Report are in Appendix~\ref{appendix:experiment_1}.

\paragraph{(1) Performance Scores} We obtain two main performance metrics through the interview process:
\begin{itemize}[nosep, topsep=0pt, itemsep=0pt]
    \item \texttt{\textbf{Problem Solving Ability}} – 
    Measures how effectively the model solves a given problem at its \(n\)-th interaction with the \Interviewer{}. We define \(\textrm{Score}_{\textrm{seed}}\text{@}n\) as the model's performance at the \(n\)-th interaction, where an interaction involves receiving feedback and making revisions. Thus, \(\textrm{Score}_{\textrm{seed}}\text{@}1\) corresponds to the LLM-as-a-Judge setting without feedback, while \(\textrm{Score}_{\textrm{seed}}\text{@}n\) represents performance after \(n-1\) iterations of feedback and revision.


    \vspace{-1mm}
    \item \texttt{\textbf{Follow-up Question Handling}} 
    – Evaluates how well the model responds to follow-up questions. To ensure fairness across models with varying performance levels, we categorize follow-up question accuracy into three types: accuracy within correctly answered questions, accuracy within incorrectly answered questions, and overall accuracy.

\end{itemize}

These metrics reflect abilities crucial in real-world interactions with LLMs, where responding effectively to feedback and follow-up questions significantly influences user satisfaction beyond just answering the initial query. 
\begin{table*}[t]
\small
\centering
\renewcommand{\arraystretch}{1.2} 
\begin{tabular}{@{}lp{0.7\textwidth}@{}}
\specialrule{1.2pt}{0pt}{3pt} 
\textbf{Follow-Up Question} & \textbf{\textsc{{\underline{Type}}}} and Example \\
\specialrule{1.2pt}{3pt}{3pt} 
\quad {Knowledge/Factuality} & \textbf{\textsc{\underline{{Additional Facts}}}} \quad \textbf{Can you provide an example} of determining the 6th roots of unity and specifying their arguments in radians? \\
\midrule
\quad \text{Reasoning} & \textsc{\underline{\textbf{{Clarification}}}} \quad \textbf{What does it mean} for \(a\) and \(b\) to be consecutive integers? \\

\cmidrule{2-2}
 & \textbf{\textsc{\underline{{Rationale}}}} \quad \textbf{How did you determine} Jasmine’s rate of water consumption per mile? \\ 
\midrule 
\quad \text{Instruction Following} & \textsc{\underline{\textbf{{Modification of Conditions}}}} \quad \textbf{If we change the previous question} and assume that each sister of David has two brothers, how many brothers would David have? \\
\cmidrule{2-2}
 & \textsc{\underline{\textbf{{Additional Explanation}}}} \quad \textbf{Can you provide a specific example of how you would address} a potential objection your friend might have about public speaking in your persuasive email? \\
\cmidrule{2-2}
 & \textsc{\underline{\textbf{{Correction}}}} \quad \textbf{Can you now reformulate} your earlier reply, outputting it in JSON format and only including books published after 1980? \\
\specialrule{1.2pt}{3pt}{0pt} 
\end{tabular}
\caption{\textbf{Examples of Follow-Up Questions posed by the interviewer (GPT-4o)} along with the tasks and types.}
\label{table:real_ex_feedbac_followup}
\end{table*}

\paragraph{(2) Error Analysis and Examples} A detailed breakdown of error types across multiple interactions, along with illustrative examples. These highlight common failure reasons and examples in the models, offering insights into their limitations and areas for improvement.
\paragraph{(3) Comprehensive Summary}
A summary of the model's behavior throughout the interview process across all samples. It highlights the model's strengths and weaknesses in performing the task. (e.g. GPT-4o in DepthQA: \texttt{``The model struggles with conciseness, often providing overly detailed responses...''})

\section{Experimental Setup}
\label{sec:exp_setup}
In this section, we describe the overall experimental setup used in \S~\ref{sec:benchmarking}-~\ref{sec:DataContam}, as well as the design of the \textbf{LLM-as-an-Interviewer} framework.
In our experiment, we use three tasks: {Reasoning, Knowledge, and Instruction Following}. 
For {Reasoning}, we use {MATH}~\cite{MATH} and the reasoning subset of {MINT}~\cite{wang2023mint}.  
For {Knowledge}, we use {DepthQA}~\cite{depthqa}, a dataset of real-world STEM-related questions assessing factuality, rebuilt from the TutorEval dataset.  
For {Instruction Following}, we leverage {MT-Bench}~\cite{chiang2024chatbotarenaopenplatform}, which evaluates an ability to follow multi-turn instructions.

\subsection{Models}
We list all the models\footnote{We use the Azure API for GPT models, leveraging the GPT-4o-0514 and GPT-3.5-turbo-0125 versions.}\footnote{We use the \href{https://api.together.xyz/}{Together API} to access Llama models.
} used throughout our experiments in Section~\ref{sec:benchmarking}-\ref{sec:DataContam}.
\paragraph{\Interviewer{} Models} 
We adopt GPT-4o as the default Interviewer, as it demonstrates the best performance (about 90\% accuracy as an Interviewer) in our analyses.
In Appendix~\ref{appendix:experiment_2}-\ref{appendix:experiment_3}, we present detailed analyses on the accuracy of various interviewers, error analysis, and average cost.
\paragraph{\Interviewee{} Models} 
We use a total of 8 models for the Interviewee, including 2 proprietary models (GPT-4o~\cite{openai2024gpt4ocard}, GPT-3.5) and 6 open-source models (Llama-3.1-\{8,70\}b~\cite{llama3}, deepseek-Math-7b~\cite{deepseekmath}, Qwen2.5-Math-7b~\cite{qwenmath}, OLMoE~\citep{muennighoff2024olmoeopenmixtureofexpertslanguage} and Zephyr-7b~\citep{tunstall2023zephyrdirectdistillationlm}).
\subsection{Interview Process Design}
LLM-as-an-Interviewer framework adapts to each task type's characteristics while maintaining rigorous evaluation standards. 
This section describes the interview process design used in the experiments. Additional details regarding the prompts used are provided in the {Appendix~\ref{appendix:experiment}-\ref{appendix:prompt}}. An example of the full interview log is attached in the Appendix~\ref{appendix:full_log}.

\begin{table*}[t]
\centering
\small
\begin{tabular}{>{\raggedright\arraybackslash}p{2.5cm}|ccccc|ccccc}
\toprule
 & \multicolumn{5}{c|}{\textbf{Reasoning (MATH)}} & \multicolumn{5}{c}{\textbf{Factuality (DepthQA)}} \\
\cmidrule(l){2-11}
\textbf{Model} & \textbf{Judge} & \textbf{@1} & \textbf{@2} & \textbf{@3} & \textbf{$\Delta$} & \textbf{Judge} & \textbf{@1} & \textbf{@2} & \textbf{@3} & \textbf{$\Delta$} \\%
\midrule
\textbf{GPT-4o} & 0.553 & 0.444 & 0.566 & 0.636 & 0.192 & 0.989 & 0.987 & 0.997 & 0.999 & 0.012 \\%
\textbf{GPT-3.5-turbo} & 0.162 & 0.082 & 0.236 & 0.346 & 0.264 & 0.986 & 0.916 & 0.975 & 0.984 & 0.068 \\%
\textbf{Llama3.1-70b} & 0.398 & 0.254 & 0.436 & 0.552 & 0.298 & 0.990 & 0.959 & 0.985 & 0.989 & 0.030\\%
\textbf{{Llama3.1-8B}} & 0.222 & 0.128 & 0.217 & 0.300 & 0.172 & 0.984 & 0.906 & 0.970 & 0.977 & 0.071 \\%
\textbf{DeepSeek-math} & 0.140 & 0.088 & 0.222 & 0.348 & 0.260 & 0.977 & 0.875 & 0.938 & 0.958 & 0.083 \\%
\textbf{Qwen-math} & 0.454 & 0.303 & 0.429 & 0.483 & 0.180 & 0.814 & 0.878 & 0.898 & 0.907 & 0.029 \\%
\midrule
\textbf{\textit{std.}} & \textit{0.143} & \textit{0.161} & \textit{0.123} & \textit{0.126} & \textit{-} & \textit{0.0597} & \textit{0.0485} & \textit{0.0482} & \textit{0.0481} & \textit{-} \\
\bottomrule
\end{tabular}
\caption{MATH Problem Solving Accuracy and DepthQA Precision along with Delta ($\Delta$) for six models. @k indicates the performance of the k-th interaction. Judge indicate scores of LLM-as-a-judge.}
\label{tab:benchmarking}
\end{table*}

\paragraph{Feedback Criteria} 
The \Interviewer{} provides feedback when there is an error in the model's response or final answer (if the task involves absolute answers). We prompt the \Interviewer{} to guide the model to recognize its mistakes and revise accordingly, but not to provide the correct answer directly. We provide feedback examples in Table~\ref{table:real_ex_feedbac} of the Appendix, where the \Interviewer{} gives feedback on incorrect responses or missing aspects based on the evaluation criteria.

\paragraph{Follow-Up question Type} 
For each task, we categorize follow-up question types as shown in Table~\ref{table:real_ex_feedbac_followup}.  
For Knowledge/Factuality tasks, follow-ups probe for additional facts, partially covering the role of a recall metric by identifying missing information from the model’s previous response-a common limitation in factuality evaluation~\cite{min-etal-2023-factscore}.  
For reasoning tasks, follow-ups aim to uncover the underlying rationale or failure reasons. For instruction following, we define three follow-up types based on the type of second-turn questions by users summarized in MT-Bench-101~\cite{bai-etal-2024-mt}.



\section{Evaluating LLMs with \textsc{LLM-as-an-Interviewer}}
\label{sec:benchmarking}
We simulate the interview process with various \Interviewee{} models.
We present insights from interviews, focusing on the seed question-solving phase in \S~\ref{sec:benchmarking_feedback}, and discuss the follow-up questions phase in \S~\ref{sec:benchmarking_followup}.
\paragraph{Test set} We sample 500 queries\footnote{We provide a statistical analysis of the sampling in Appendix~\ref{appendix:stats-sampling}.} from both MATH and DepthQA, ensuring an equal distribution across the difficulty levels specified in the datasets. We use all 316 samples from the MINT reasoning set and 80 samples from MT-Bench.

\paragraph{\Interviewee{} Models} We use 6 models for the Interviewee, GPT-4o, GPT-3.5, Llama-3.1-\{8,70\}b, DeepSeek-Math-7b, Qwen2.5-Math-7b. 
\subsection{Seed Question Solving Phase}
Table~\ref{tab:benchmarking} presents the scores of seed questions in MATH and DepthQA for both the LLM-as-a-Judge and LLM-as-an-Interviewer settings. In this experiment, query modification is applied at the beginning of the interview, and the same questions are used for the interviewee models. We present the results from MINT in Table~\ref{tab:mint_result} of Appendix~\ref{appendix:simulate_mint}.
\paragraph{Comparison with LLM-as-a-Judge Score}
There is a clear performance drop in the LLM-as-an-Interviewer setting, with most models struggling more when handling modified queries. In MATH, most models achieve performance comparable to the Judge during the second interaction. For DepthQA, all models except GPT-4o experience a significant performance decline, while GPT-4o shows only a minimal decrease of 0.2\%. In addition, the evaluation criteria that assess both the reasoning process and the final answers can contribute to this performance drop. For example, GPT-3.5 correctly answers the questions but introduces errors in the solving process in 3\% of the MATH test set.

\paragraph{Model's Capability to Revise Through Feedback}
\label{sec:benchmarking_feedback}
Across all models, performance improves when feedback is provided for incorrect answers ($\text{@}2, \text{@}3$), compared to the initial accuracy ($\text{@}1$). Interestingly, as the models receive feedback and iteratively revise their answers, the standard deviation (\textit{std}) of scores across models generally decreases compared to that in the LLM-as-a-Judge setting and @1 score. This is mainly because models that initially perform poorly show substantial improvement through interaction.


\subsection{Follow-UP Question Solving Phase}
\label{sec:benchmarking_followup}

\begin{table*}[t]
\centering
\small
\begin{tabular}{@{}ll|ccc|ccc|ccc@{}}
\toprule
\multicolumn{2}{l|}{}                           & \multicolumn{3}{c|}{\textbf{Reasoning (MATH)}}                          & \multicolumn{3}{c|}{\textbf{Reasoning (MINT)}}   & \multicolumn{3}{c}{\textbf{Factuality (DepthQA)}}                      \\ \cmidrule(l){3-11} 
\textbf{Model}                                  & \multicolumn{1}{c|}{} & Success   & Fail  & Overall  
                                                & Success   & Fail  & Overall  
                                                & Success   & Fail  & Overall  \\ \midrule
\textbf{GPT-4o}                                 &                       & 0.99 & 0.17 & 0.93 & 0.96 & 0.82 & 0.94 & 0.93   & 0.50   & 0.92 \\ 
\textbf{GPT-3.5}                                &                       & 0.90 & 0.56 & 0.82 & 0.85 & 0.70 & 0.80 & 0.73   & 0.64   & 0.73 \\ 
\textbf{Llama3.1-70b}                           &                       & 0.91 & 0.22 & 0.84 & 0.91 & 0.74 & 0.87 & 0.85   & 0.63   & 0.83 \\ 
\textbf{Llama3.1-8B}                            &                       & 0.89 & 0.07 & 0.76 & 0.75 & 0.65 & 0.72 & 0.80   & 1.00   & 0.80 \\ 
\textbf{DeepSeek-math}                          &                       & 0.82 & 0.38 & 0.69 & 0.42 & 0.41 & 0.42 & 0.68   & 0.50   & 0.65 \\ 
\textbf{Qwen-2.5-math}                          &                       & 0.90 & 0.08 & 0.78 & 0.55 & 0.34 & 0.50 & 0.61   & 0.32   & 0.52 \\ \bottomrule
\end{tabular}
\caption{Follow-up accuracy of six models on MATH, MINT, and Depth-QA datasets.}
\label{tab:followup}
\end{table*}

Table~\ref{tab:followup} presents follow-up question accuracy in Reasoning and Factuality tasks. Overall, models exhibit higher accuracy on follow-up questions for correctly answered seed questions than for failed ones. In
\S~\ref{sec:mtbench}, we provide a discussion based on the MT-Bench evaluation.

\subsubsection{Comparing with Static Multi-turn Benchmark}
\label{sec:mtbench}

We compare follow-up questions in LLM-as-an-Interviewer with MT-Bench~\cite{chiang2024chatbotarenaopenplatform}, a 2-turn static benchmark with 80 samples across various tasks (e.g., writing, math, coding, roleplay).

MT-Bench primarily uses general second-turn questions that are independent of previous responses (e.g., \textit{"Explain like I'm five,"} \textit{"Change format"}), limiting  diversity, and realism. In contrast, LLM-as-an-Interviewer generates context-aware follow-ups (e.g., \textit{"You mentioned Thomas might be in a clinical trial. Can you explain what a clinical trial is and why it requires daily hospital visits?"}).  
Some MT-Bench tasks reference previous responses (e.g., \textit{"Explain the N-th example from your previous response."}), which can cause contradictions if the model fails to follow instructions in turn 1, as seen in the following example.
\begin{tcolorbox}[
    colback=white, 
    colframe=gray!50!white, 
    title={\fontsize{9}{9}\selectfont \textbf{MT-bench Failure Example of Deepseek}}, 
    fonttitle=\color{black}, 
    breakable
]
\fontsize{8}{9}\selectfont 
\textbf{Question-Turn-1:} \\
Help me construct a catchy, yet scientifically accurate, \textbf{headline for an article} on the (...). Propose 4 options. \\
\textbf{Model Response (Deepseek):} \textcolor{red}{(Irrelevant Response)}

\texttt{from sympy import symbols, simplify, solve\\
def solve\_equation():\\
    x = symbols('x')\\
    equation = x**2 - 16 (...)\\}

\textbf{Question-Turn-2:} \\
Alter your previous response. \hllow{Make the following adjustments to the 2nd option}: 1. Make the tone sound casual (...)

\end{tcolorbox}

\begin{table}[t]
    \centering
    \small
    \begin{tabular}{lcc|cc}
        \toprule
        & \multicolumn{2}{c|}{\textbf{MT-bench}} & \multicolumn{2}{c}{\textbf{Interviewer}} \\
        & & & \textbf{GPT-4o} & \textbf{Qwen} \\
        \midrule
        \textbf{Dependent} & 27 &  & \textbf{54 }& \textbf{44 }\\
        \textbf{Independent} & \textbf{53} &  & 26 & 36 \\
        \bottomrule
    \end{tabular}
    \caption{Dependent and Independent question Counts in 2-turn questions. GPT-4o and Qwen denote the \Interviewee{}.}
    \vspace{-5mm}
    \label{tab:dependent_independent}
\end{table}

Table~\ref{tab:dependent_independent} shows the number of dependent and independent follow-up questions for each setting. 
For the Interviewer setting, we compare GPT-4o, the best-performing model at turn-1, and Qwen, the worst-performing model with low instruction-following ability\footnote{See Table~\ref{table:mtbench_results} in Appendix~\ref{appendix:simulate} for the results in MT-bench}. 
Aligned with MT-Bench failure examples, Qwen receives more independent follow-ups (e.g., topic shifts), while GPT-4o receives more dependent ones. 
As such, considering previous responses dynamically during multi-turn evaluations is crucial for assessing the model in a more diverse and natural context.

\section{Can \textsc{LLM-as-an-Interviewer} Mitigate Data Contamination Issue?} 
\label{sec:DataContam}

This section demonstrates the efficiency of the modification strategy in generating questions based on established benchmarks.
\subsection{Experimental Setting}

We establish various settings depending on the model's training dataset. Specifically, we use two open-source models, OLMoE and Zephyr-7b, whose training datasets are disclosed, and fine-tune them on different training sets. These settings allow us to evaluate both \textbf{Uncontaminated Models}—which do not access the test set during training, and \textbf{Contaminated Models}—which are trained with exposure to the test set.

\paragraph{Uncontaminated Model Training}  
As shown in the setting column of Table~\ref{tab:math-contam}, we train uncontaminated models using three distinct configurations:  
\begin{itemize}[nosep, topsep=0pt, itemsep=0pt]
    \item[1] Models trained exclusively on the train set of the target dataset (In-Distribution, ${Train}_{ID}$).
    \item[2] Models trained on both the ${Train}_{ID}$ and additional datasets from the same domain (Out-Of-Distribution, ${Train}_{OOD}$).
    \item[3] Models trained solely on ${Train}_{OOD}$.
\end{itemize}

\paragraph{Contaminated Model Training}  
For contaminated models, we employ four different configurations:  
\begin{itemize}[nosep, topsep=0pt, itemsep=0pt]
\item[4] Models trained exclusively on the test set of the target dataset (In-Distribution, ${Test}_{ID}$).
\item[5] Models trained on both the ${Test}_{ID}$ and ${Train}_{ID}$ sets.
\item[6] Models trained on the ${Test}_{ID}$ combined with an instruction-tuning (${Instruct}$) dataset.
\item[7] Models trained on both ${Test}_{ID}$ and ${Train}_{OOD}$.
\end{itemize}

\paragraph{Dataset}  
We use the in-distribution test set(${Test}_{ID}$) when testing the model, and MATH and DepthQA are used as in-distribution datasets. Since DepthQA is not pre-divided into train and test sets, we manually split it into 851 instances for the $\text{DepthQA}_\text{train}$ set and 848 instances for the $\text{DepthQA}_\text{test}$ set. For other datasets, we randomly select 2000 samples each. For the out-of-distribution dataset of MATH, we use GSM8K. 
For MATH, we experiment with all seven settings, while for DepthQA, we include only the four settings that do not involve out-of-distribution (OOD) data.

Details about model training, including training configurations, parameters, and datasets used, are provided in Appendix~\ref{appendix:contam_1}.


\definecolor{green1}{HTML}{E3F9E5}  
\definecolor{green2}{HTML}{A5D6A7}  
\definecolor{green3}{HTML}{66BB6A}  

\begin{table}[t]
\centering
\small
\begin{tabular}{@{}ll|cccc@{}}
\toprule
\multicolumn{2}{l|}{}                                            & \multicolumn{2}{c|}{\textbf{OLMoE}}                                                                   & \multicolumn{2}{c}{\textbf{Zephyr}}                                                               \\ \cmidrule(l){3-6} 
\textit{Model.}                          & \multicolumn{1}{c|}{} & \multicolumn{1}{c|}{Judge.}              & \multicolumn{1}{c|}{Interv.}                                   & \multicolumn{1}{c|}{Judge.}              & Interv.                                    \\ \midrule
\textit{\textbf{Uncontam.}}              &                       &                                         & \multicolumn{1}{c|}{}                                       &                                         &                                        \\
\hspace{3mm}$\text{Train set}_\text{ID} $           & {[}1{]}               & \cellcolor[HTML]{FFFBEF}0.10            & \multicolumn{1}{c|}{\cellcolor[HTML]{FFFEF8}0.05}               & \cellcolor[HTML]{FFFBEF}0.10            & \cellcolor[HTML]{FFFFFF}0.01               \\
\hspace{3mm}+$\text{Train set}_\text{OOD} $         & {[}2{]}               & \cellcolor[HTML]{FFFDF4}0.07            & \multicolumn{1}{c|}{\cellcolor[HTML]{FFFEF8}0.05}               & \cellcolor[HTML]{FFFCF3}0.08            & \cellcolor[HTML]{FFFEFA}0.04               \\
\hspace{3mm}$\text{Train set}_\text{OOD} $          & {[}3{]}               & \cellcolor[HTML]{FFFDF4}0.07            & \multicolumn{1}{c|}{\cellcolor[HTML]{FFFDF6}0.06}               & \cellcolor[HTML]{FFFDF6}0.06            & \cellcolor[HTML]{FFFFFE}0.02               \\ \midrule
\textbf{Avg.}                              &                       & 0.08            & \multicolumn{1}{c|}{0.05}               & 0.08               & 0.02            \\ \midrule

\textit{\textbf{Contam.}}                &                       &                                         & \multicolumn{1}{c|}{}                                       &                                         &                                         \\
\hspace{3mm}$\text{Test set}_\text{ID} $            & {[}4{]}               & \cellcolor[HTML]{FFD666}{0.84}   & \multicolumn{1}{c|}{\cellcolor[HTML]{FFFCF3}{0.08}}      & \cellcolor[HTML]{FFE291}0.61            & \cellcolor[HTML]{FFFBEF}0.10               \\
\hspace{3mm}+$\text{Train set}_\text{ID} $          & {[}5{]}               & \cellcolor[HTML]{FFDA73}0.77            & \multicolumn{1}{c|}{\cellcolor[HTML]{FFF9E6}0.15}               & \cellcolor[HTML]{FFE08B}0.64            & \cellcolor[HTML]{FFF8E2}0.17               \\
\hspace{3mm}+Instruct                    & {[}6{]}               & \cellcolor[HTML]{FFE49A}0.56            & \multicolumn{1}{c|}{\cellcolor[HTML]{FFFAEB}0.12}               & \cellcolor[HTML]{FFF7DE}0.19            & \cellcolor[HTML]{FFFCF1}0.09               \\
\hspace{3mm}+$\text{Train set}_\text{OOD} $         & {[}7{]}               & \cellcolor[HTML]{FFDB77}0.75            & \multicolumn{1}{c|}{\cellcolor[HTML]{FFFBEF}0.10}               & \cellcolor[HTML]{FFEBB4}0.42            & \cellcolor[HTML]{FFFCF3}0.08               \\ \midrule
\textbf{Avg.}                    &                       & 0.69            & 0.12               & 0.42               & 0.11            \\ \bottomrule
\end{tabular}
\caption{\textbf{Performance of uncontaminated and contaminated models on MATH dataset}. "Judge" refers to the LLM-as-a-Judge setting, and "Interv" refers to the LLM-as-an-Interver setting.}
\label{tab:math-contam}
\end{table}

\subsection{Result}
\label{sec:contam_acc}
Table \ref{tab:math-contam} presents the effectiveness of the modifying strategy in mitigating data contamination, which is applied while the \Interviewer{} prepares the seed queries. The tables compare accuracy for both the original queries just as LLM-as-an-Judge setting (\textit{Judge}) and modified queries (\textit{Interview}). Settings [1, 2, 3] represent uncontaminated models, serving as a proxy for the model's true ability on the task. The goal of the strategy is to bring the performance of contaminated models closer to these uncontaminated models.

In MATH, contaminated models often outperform uncontaminated ones under the \textit{Judge} setting. These contaminated models achieve significantly higher accuracy, averaging 0.7 for OLMoE and 0.4 for Zephyr. However, applying the modification strategy leads to a substantial performance drop, making their performance comparable to the uncontaminated model. A similar trend is observed in the DepthQA results (Table~\ref{tab:depthqa-contam} in Appendix~\ref{appendix:contam}). The results show that our framework mitigates the contamination issue by having the model solve similar yet modified questions.



\section{Discussion}
\label{sec:reliable}
\label{sec:reliability-bias}
We assess whether LLM-as-an-Interviewer exhibits verbosity bias and maintains robustness across multiple runs, a common consideration in model-based evaluation~\cite{zheng2023judgingllmasajudgemtbenchchatbot}.\footnote{We also provide an analysis of self-enhancement bias in the appendix~\ref{appendix:Reliability_2}.}
\paragraph{Verbosity Bias} 
To assess whether the framework favors longer answers, we examine the correlation between answer length and score using the Depth-QA task. GPT-4o acts as the \Interviewer{}, with six models as \Interviewee{}.\footnote{\Interviewee{}: GPT-4o, GPT-3.5-turbo, Llama-3.1-70b, Llama-3.1-8b, DeepSeek-Math, Qwen2-Math.}

Figures~\ref{fig:verbosity} in the Appendix~\ref{appendix:Reliability_1} reveal that the linear correlation (\textit{r}) between length and score weakens as interactions increase. In the LLM-as-a-Judge setting , a significant positive correlation is found (\textit{r = .371, p < .05}); however, as interactions progress, the \textit{p}-value rises and \textit{r}-value declines. This indicates that our interview process could reduce verbosity bias and length dependency, alongside the length control mechanism for adjusting win rates or scores~\cite{dubois2024lengthcontrolledalpacaevalsimpleway}.

\begin{table}[t]
\small
\centering
\begin{tabular}{@{}lccccc@{}}
\toprule
\textbf{Interact.} & \textbf{@1}   & \textbf{@2}   & \textbf{@3}   & \textbf{@4}   & \textbf{@5}   \\ \midrule
\textit{DepthQA}      & 0.0244        & 0.0151        & 0.0116        & 0.0090        & 0.0082        \\ \midrule   
\textit{MATH}         & 0.0358        & 0.0480        & 0.0583        & 0.0642        & 0.0704   \\
\textit{\hspace{3mm}easy}    & 0.1115        & 0.1127        & 0.0839        & 0.0797        & 0.0890        \\
\textit{\hspace{3mm}hard}    & 0.0531        & 0.1032        & 0.1281        & 0.1372        & 0.1332        \\ \bottomrule
\end{tabular}
\caption{Standard Deviation of Multiple Runs Across Interactions.}
\vspace{-3mm}
\label{tab:interactions}
\end{table}
\paragraph{Does \textsc{LLM-as-an-Interviewer} Function Robust?}
\label{sec:reliability-robust}
We assess the robustness of our framework across multiple interactions by repeating the experiment five times for each setting. Specifically, We calculate the standard deviation of these five runs for each combination of different \Interviewer{} temperatures~(0 or 1) and different \Interviewee{} models. The \Interviewee{}’s temperature remains fixed at 1. We use GPT-4o as the \Interviewer{} and GPT-3.5 and Llama-3.1-\{8/70\}b as the \Interviewee{}.

As shown in Table~\ref{tab:interactions}, the standard deviation for DepthQA decreases with more interactions. In contrast, for MATH, it increases over time. Breaking MATH down by difficulty, we find that for easier problems (levels ≤2, MATH-easy), the standard deviation remains stable or decreases, whereas for harder problems (levels ≥3, MATH-hard), it rises due to initially low accuracy and occasional correct answers from feedback.
Excluding MATH-hard, the standard deviation remains stable, demonstrating the framework’s robustness in most cases.

\section{Conclusion}

In this paper, we introduce LLM-as-an-Interviewer, a novel approach for evaluating LLMs through interviews. Our multi-turn evaluation framework with Interview Report reveals the abilities of LLMs, such as revising answers based on feedback or handling follow-up questions, while providing comprehensive insights into the task. Our analysis demonstrates that our framework mitigates common issues in static evaluation methods, including verbosity bias and data contamination. We hope our framework is widely adopted and brings a paradigm shift in LLM evaluation.
\section{Limitations}
\label{appendix:limitation}

While our framework addresses several biases and limitations of the LLM-as-a-judge approach, it still inherits some of the inherent limitations of large language models, such as inconsistency and the potential to introduce additional biases, similar to the issues raised in the LLM-as-a-judge approach.
\section*{Acknowledgements}
This research project has benefitted from the Microsoft Accelerate Foundation Models Research (AFMR) grant program through which leading foundation models hosted by Microsoft Azure along with access to Azure credits were provided to conduct the research.
This work was supported by Institute for Information \& communications Technology Promotion(IITP) grant funded by the Korea government (MSIP) (No. RS-2024-00443251, Accurate and Safe Multimodal, Multilingual Personalized AI Tutors)
\bibliography{anthology,custom}
\bibliographystyle{acl_natbib}
\newpage
\newpage

\appendix

\newpage
\newpage
\clearpage

\section{AI Writing Assistance}
We used ChatGPT for grammar and language refinement of the manuscript.
\section{LLM-as-an-Interviewer}\label{appendix:Method}
\subsection{Implementation Details}
\label{appendix:Method_1}
Building upon the interview framework described in \S~\ref{sec:interview_process}, this section explains our interview system implementation. The system operates as a two-agent dialogue with additional complexity in controlling the \Interviewer{} to conduct evaluations accurately and efficiently. Controlling Interviewer is done in two stages: (1) keeping the interview state in track, and (2) using pre-written prompts for each state. While we implemented the interview process specifically for MATH and STEM domains, the framework can be easily adapted to other domains with minor modifications.

\subsection{Adapting to New Tasks}
\label{appendix:Method_2}
To adapt a new task, one can modify the Main Interview process, specifically the response evaluation criteria and follow-up question generation strategy. This can be done by rewriting the pre-written prompts. Here are examples of adaptations for different domains:

\paragraph{Programming Interviews:}
\begin{itemize}
    \item \textbf{Response Evaluation:} Code correctness, efficiency, and style
    \item \textbf{Follow-up Generation:} Edge cases, optimization opportunities, or alternative implementations
\end{itemize}

\paragraph{Language Proficiency:}
\begin{itemize}
    \item \textbf{Response Evaluation:} Grammar accuracy, vocabulary usage, and fluency
    \item \textbf{Follow-up Generation:} Complex linguistic scenarios or cultural context questions
\end{itemize}

\paragraph{Customer Service Evaluation}
\begin{itemize}
    \item \textbf{Response Evaluation:} Empathy in customer interactions, response tone and professionalism, policy adherence
    \end{itemize}
    
    \textbf{Follow-up Generation:} 
    \begin{itemize}
        \item For acceptable responses: Escalate scenario complexity (e.g., adding multiple complaints or time pressure), Rationale of responses
        \item For problematic responses: Probe understanding of customer service principles or store policies
    \end{itemize}
    
    
    

\subsection{Algorithmic Implementation}
\label{appendix:Method_0}
\begin{algorithm*}
\caption{Interview Process} \label{interview_algorithm}
\begin{algorithmic}[1] 
\Require $Q_{\text{init}}$: Initial question
\Require $\text{Modify}(\cdot)$: Function to modify the initial question
\Require $I$: Interviewer
\Require $i$: Interviewee
\Require $G$: Grader
\Require $AskFollowup(\cdot)$: Function to ask a follow-up question
\Require $MAX\_QUESTIONS$: Maximum number of questions
\Require $MAX\_RETRIES$: Maximum number of retries for each question

\Statex \textbf{Constants:}
\State $\text{MAX\_QUESTIONS} \gets N$ \Comment{Maximum number of questions}
\State $\text{MAX\_RETRIES} \gets M$ \Comment{Maximum retries per question}

\Statex \textbf{Variables:}
\State $Q_{\text{modified}} \gets \text{Modify}(Q_{\text{init}})$ \Comment{Modified list of seed questions}
\State $\text{feedbacks} \gets \text{[]}$
\State $\text{questions\_count} \gets 0$
\State $\text{termination} \gets \text{False}$

\Statex \textbf{Execution:}
\For{each question $q_{\text{seed}} \in Q_{\text{modified}}$}
    \State $\text{questions\_count} \gets \text{questions\_count} + 1$
    \State $q \gets q_{\text{seed}}$ \Comment{Start with the seed question}

    \While{\textbf{not} $\text{termination}$}
        \State $response \gets i(q)$ \Comment{Interviewee answers the question}
        \State $\text{feedback}, \text{is\_correct} \gets G(q, response)$ \Comment{Grader evaluates response}
        \State $\text{feedbacks.append(feedback)}$

        \If{$\text{is\_correct}$}
            \State \textbf{break} \Comment{Move to the next question if correct}
        \EndIf

        \Statex \hspace{2em} \textbf{Retries for Incorrect Answers:}
        \State $\text{retries} \gets 0$
        \While{\textbf{not} $\text{is\_correct}$ \textbf{and} $\text{retries} < \text{MAX\_RETRIES}$}
            \State $\text{hint} \gets \text{GetHint}(q, response, \text{feedback})$ \Comment{Generate hint}
            \State $response \gets i(hint)$ \Comment{Interviewee retries the question}
            \State $\text{feedback}, \text{is\_correct} \gets G(q, response)$
            \State $\text{feedbacks.append(feedback)}$
            \State $\text{retries} \gets \text{retries} + 1$
        \EndWhile

        \Statex \hspace{2em} \textbf{Follow-up Question:}
        \If{$\text{questions\_count} > \text{MAX\_QUESTIONS}$}
            \State $\text{termination} \gets \text{True}$ \Comment{Terminate interview after max questions}
        \EndIf
        \State $q \gets \text{AskFollowup}(q)$ \Comment{Switch to a follow-up question}

    \EndWhile
\EndFor

\end{algorithmic}
\end{algorithm*}

Algorithm~\ref{interview_algorithm} presents the algorithmic implementation of the interview process.
\clearpage
\section{Experimental Setup}
\label{appendix:experiment}
\subsection{Experimental Details}
\label{appendix:experiment_1}
\paragraph{Preparing Seed Question:}\Interviewer{} model generates modified versions of seed questions through unclarification or paraphrasing

For Arithmetic Reasoning tasks (MATH), LLM Interviewer replaces numerical values in a query with unknown variables (e.g., \textit{x}) or omits specific details. For example, \textit{"If there were 5 chocolates and you ate \textbf{1}, how many are left?"} becomes \textit{"If there were 5 chocolates and you ate \textbf{x}, how many are left?"}. This maintains the same reasoning level while preventing the direct application of memorized solutions, as seen in methods like those of \citet{mirzadeh2024gsmsymbolicunderstandinglimitationsmathematical}. For factuality-based long-form tasks (depth-QA), new problems are created by referencing ground truth solutions and questions from the original dataset to ensure relevance and diversity. Examples of these strategies are in Table~\ref{tab:preinterview-ex}.

\paragraph{Interview Process:} Consists of two primary components:
\begin{enumerate}
    \item \textbf{Response Evaluation:} \Interviewer{} assesses \Interviewee{} responses using domain-specific criteria (mathematical correctness for MATH; factual accuracy and completeness for STEM) and identifies error sources.
    
    \item \textbf{Follow-up Question Generation:} Based on evaluation results, \Interviewer{} either probes deeper concepts (for correct responses) or provides targeted feedback to reveal misconceptions (for incorrect responses). 
\end{enumerate}
\paragraph{Evaluation Metric:}For MATH, our evaluation framework implements binary correctness assessment for both final answers and solution processes, along with error categorization that distinguishes between conceptual understanding, misinterpretation, and calculation errors. The framework also incorporates step-by-step verification of mathematical reasoning and generates targeted follow-up questions when errors are detected. 

For DepthQA~\cite{depthqa}, our evaluation framework uses FActScore~\cite{min-etal-2023-factscore} as a metric to calculate factual precision in the model's long-form generation.
These follow-up questions partially address a limitation of previous fact-based evaluations by enabling a somewhat broader assessment that includes aspects of recall in addition to precision.
employs a more nuanced approach, decomposing responses into atomic facts for systematic assessment. 
The framework also evaluates quality across multiple dimensions, including completeness, redundancy, readability, and depth, while generating follow-up questions to probe understanding of missing concepts.
\paragraph{Interview Report:} The interview report, based on the detailed interaction logs from the Main Interview stage, is shown to the user at the end of the interview. This content for the report is explained in \S~\ref{sec:interview_report}. Followings are the example of Interview Report. 
\subsection{Statistical analysis of the Sampling}
\label{appendix:stats-sampling}
We compare the sampling variability of our framework using bootstrapping (sampling with replacement). From the 500 samples, we randomly sample \(n\) samples, repeated the process 10,000 times, and calculated the p-value. The results showed no significant statistical difference between \(n \geq 100\) samples and 500 samples, with results comparable to those of LLM-as-a-Judge.

\begin{table}[h]
\centering
\small
\begin{tabular}{c|ccc}
\hline
\textbf{Number of Samples } & 100 & 300 & 500 \\ \hline
\textbf{Interviewer}  & p = 0.809 & p = 0.832 & p = 0.766 \\ \hline
\textbf{Judge}         & p = 0.464 & p = 0.740 & p = 0.714 \\ \hline
\end{tabular}
\caption{Statistical Comparison of Sampling Variability}
\end{table}

\newpage
\begin{tcolorbox}[colback=white, colframe=yellow!50!yellow, title=MATH{,} Interviewee: GPT-4o\\(Interviewer: GPT-4o),fonttitle=\color{black},breakable]

\textbf{1. Performance Scores(\%)}
\begin{itemize}
  \item \textbf{Accuracy(@ \# of interaction):} \\ 72(@1), 82(@2), 84(@3)
  \item \textbf{Follow-up Accuracy:}
    \begin{itemize}
\item Total 93
      \item Rationale Questions 99
      \item Clarification  17
  \end{itemize}

\end{itemize}

\textbf{2. Error Types \& Examples}
\\{*} Frequent Rate for Each Type
\\{*} Examples are omitted
\begin{itemize}
  \item Misinterpret.: 0.17
  \item Calculation: 0.26 
  \item Conceptual: 0.16 
\end{itemize}

\textbf{3. Summary} \\ 
The model demonstrates strong mathematical problem-solving abilities, particularly in algebra, calculus, and logical reasoning. It excels in providing detailed, step-by-step explanations and is effective in handling follow-up questions and user feedback. However, \textbf{it struggles with interpreting geometric properties}, maintaining accuracy in complex calculations, and integrating specific\textbf{ user-provided details without explicit prompts}. The model often requires user intervention to correct initial assumptions and may provide overly verbose explanations. Improvements in handling ambiguous or incomplete information and better error-checking mechanisms could enhance its effectiveness.
\end{tcolorbox}

\begin{tcolorbox}[colback=white, colframe=yellow!50!yellow, title=MATH{,} Interviewee: Llama-3.1-8b\\(Interviewer: GPT-4o),fonttitle=\color{black}]
\textbf{1. Performance Scores(\%)}
\begin{itemize}	
  \item \textbf{Accuracy(@ \# of interaction):} \\ 58(@1), 68(@2), 75(@3)
  \item \textbf{Follow-up Accuracy:} 
  \begin{itemize}
\item Total 76
      \item Rationale Questions 89
      \item Clarification  7
  \end{itemize}

\end{itemize}

\textbf{2. Error Types \& Examples}
\\{*} Frequent Rate for Each Type
\\{*} Examples are omitted
\begin{itemize}
  \item Misinterpret.: 0.45
  \item Calculation: 0.26 
  \item Conceptual: 0.27
\end{itemize}

\textbf{3. Summary} \\ 
The model demonstrates strong mathematical problem-solving skills, particularly in algebra, quadratic equations, and logical reasoning. It excels in providing clear, step-by-step explanations, making it useful for educational purposes. The model is responsive to follow-up questions and user feedback, showing adaptability and a willingness to improve its responses. However, it struggles with\textbf{ maintaining accuracy in complex calculations, avoiding unnecessary complexity, and handling more abstract or less structured problems}. It may also require user \textbf{intervention to correct its approach and simplify its methods}. Overall, the model is effective in structured mathematical tasks but needs improvement in dealing with more complex, ambiguous, or creative problems.
\end{tcolorbox}

\newpage
\begin{tcolorbox}[colback=white, colframe=green!50!gray, title=DepthQA{,} Interviewee: GPT-4o\\(Interviewer: GPT-4o),fonttitle=\color{black},breakable]

\textbf{1. Performance Scores(\%)}
\begin{itemize}
  \item \textbf{Precision(@ \# of interaction):} \\ 96.8(@1), 98.8(@2), 99.2(@3)
  \item \textbf{Follow-up Accuracy (Additional Facts):}
    \begin{itemize}
\item Total 92
  \end{itemize}

\end{itemize}

\textbf{2. Response Quality \& Examples}
\\{*} Score for Each Type
\\{*} Examples are omitted
\begin{itemize}
  \item completeness.: 83.5
  \item redundancy: 64.9
  \item readability: 96.9
  \item depth: 64.9
\end{itemize}

\textbf{3. Summary} \\ 
The model excels in providing detailed, structured, and accurate explanations, particularly in technical, scientific, and mathematical domains. It effectively breaks down complex concepts into understandable parts and maintains coherence and continuity in responses, making it suitable for educational purposes. The model handles follow-up questions well and positively acknowledges user feedback, although it may not dynamically adapt based on feedback within a single session. \textbf{However, the model struggles with conciseness,} often providing \textbf{overly detailed responses that can overwhelm users seeking brief answers.} It may also have difficulty with highly specialized or nuanced queries, maintaining context over multiple interactions, and handling ambiguous or poorly defined questions. Additionally, the model may not always incorporate the latest information or trends beyond its training data. Overall, the model demonstrates strong capabilities in delivering comprehensive and accurate information but could improve in providing concise answers, handling more abstract or context-specific queries, and better incorporating user feedback.
\end{tcolorbox}

\begin{tcolorbox}[colback=white, colframe=green!50!gray, title=DepthQA{,} Interviewee: Llama-3.1-8b\\(Interviewer: GPT-4o),fonttitle=\color{black},breakable]

\textbf{1. Performance Scores(\%)}
\begin{itemize}
  \item \textbf{Precision(@ \# of interaction):} \\ 89.8(@1), 98.1(@2), 99.6(@3)
  \item \textbf{Follow-up Accuracy (Additional Facts):}
    \begin{itemize}
    \item Total 80
      \end{itemize}
\end{itemize}

\textbf{2. Response Quality \& Examples}
\\{*} Score for Each Type
\\{*} Examples are omitted
\begin{itemize}
  \item completeness.: 86.3
  \item redundancy: 59.1
  \item readability: 90.9
  \item depth: 81.8
\end{itemize}

\textbf{3. Summary} \\ 
The model excels in providing detailed, structured, and comprehensive explanations, particularly in scientific, technical, and mathematical fields. It effectively breaks down complex topics into understandable segments and uses examples to enhance clarity. The model handles follow-up questions well, maintaining coherence and continuity, and incorporates user feedback constructively to refine its responses. \textbf{However, the model struggles with highly specialized or niche topics, ambiguous queries}, and may initially provide overly detailed or redundant information. \textbf{It sometimes requires user prompts to provide deeper insights, }maintain conciseness, and ensure technical accuracy. Additionally, it may face challenges with real-time data, subjective questions, and practical complexities without further user guidance.
\end{tcolorbox}

\subsection{Feedback Example}
See Table~\ref{table:real_ex_feedbac}.






\subsection{Human Evaluation of LLM-as-an-Interviewer}
\label{appendix:experiment_2}

This section details the evaluation process and criteria for validating the three roles of \Interviewer{} models. Four authors conducted the human evaluation to systematically assess the effectiveness of each model in three key dimensions: (1) \textit{Query Modification}, (2) \textit{Feedback Generation}, and (3) \textit{Follow-up Question Generation}. 

In this study, we evaluate GPT-4o, Llama-3.1-70B, and Llama-3.1-8B as interviewers, while using GPT-4o, GPT-3.5, Llama-3.1-70B and Llama-3.1-8B as interviewees to simulate the interview with models of varying capabilities.
We validate at least 100 samples per dimension, resulting in over 300 samples per model. This sample size is comparable to or larger than the human-annotated datasets used in prior studies~\cite{wang2023mint, yu-etal-2024-kieval}. 
Table~\ref{tab:interviewer_performance} presents the evaluation results, showing the accuracy of each model’s responses along with the number of evaluated samples in parentheses.\footnote{Annotator reliability was validated by having two additional annotators label 100 cases. The resulting Fleiss’ kappa of 0.653 indicates substantial inter-annotator agreement.}

\subsubsection{Evaluation Criteria}

Each model’s responses were evaluated to determine whether they correctly followed the intended prompts and adhered to the reference solution. Annotators referenced both the problem statement and the reference solution throughout the evaluation process. The following criteria outline the assessment methodology:

\paragraph{Query Modification} 
For MATH tasks, we assessed whether the query retained only the key information necessary for solving the problem, avoiding unnecessary modifications that might alter the problem’s complexity. For DepthQA, we evaluated whether the query was reformulated into a well-structured and solvable question aligned with the reference solution.

\paragraph{Feedback Generation} 
We evaluated whether the feedback was contextually appropriate to the interviewee’s prior response. The feedback had to avoid contradictions with the reference solution while also refraining from directly revealing the correct answer. Annotators additionally reviewed the model’s intermediate reasoning (chain-of-thought) to ensure logical coherence.

\paragraph{Follow-up Question Generation} 
Follow-up questions were assessed based on their alignment with the problem type and intent. They needed to be relevant to the previous discussion, logically structured, and free from redundancy. For STEM-related questions, we ensured that follow-up queries were grounded in the reference solution and contributed meaningfully to the progression of the interview.

By adopting this structured evaluation framework, we ensure a comprehensive and transparent assessment of \Interviewer{} models, allowing for a clear comparison of their capabilities across different interview phases.
\subsubsection{Qualitative Analysis of LLM-as-an-Interviewer}

In this section, we present a qualitative analysis highlighting common issues encountered with the \Interviewer{} (GPT-4o) based on the annotation of results summarized in Table~10. Due to space constraints, we do not provide example excerpts for each case here; however, all actual responses will be included in the appendix.

\paragraph{Feedback Generation}  
We identify three primary failure modes:  
\begin{itemize}
    \item The \Interviewer{} incorrectly criticizes aspects that the \Interviewee{} has addressed correctly.
    \item The \Interviewer{} repeats identical feedback across multiple turns without modification.
    \item The \Interviewer{} rigidly enforces the reference solution as the only correct answer.
\end{itemize}
Given that the feedback phase is repeated $N$ times, these issues may contribute to an increased number of turns required for problem resolution.

\paragraph{Follow-up Question Generation}  
Two main issues arise during follow-up question generation:  
\begin{itemize}
    \item The \Interviewer{} asks follow-up questions about points the \Interviewee{} has not yet addressed.
    \item The \Interviewer{} focuses on minor or unnecessary details in follow-up questions.
\end{itemize}
Although these occurrences are relatively infrequent, such questions may hinder the extraction of meaningful insights into the model's behavior during the respective turns.

\subsubsection{Error Propagation in Multi-turn Interactions}
We measure multi-round accuracy, where, for example, the accuracy at turn 2 represents the percentage of cases in which the \Interviewer{} provides correct feedback across both turns 1 and 2. Table~\ref{tab:accuracy_degradation} shows the result of error propagation in multi-turn interaction. If each interaction is truly independent, we expect a compounding error rate, resulting in approximately a 9\% drop in accuracy by the second turn and an 18\% drop by the third. However, the observed degradation is much smaller because errors correlate across turns, with initial errors influencing subsequent interactions.

\begin{table}[h]
    \centering
    \begin{tabular}{l c c}
        \toprule
        & \textbf{Turn 2} & \textbf{Turn 3} \\
        \midrule
        Accuracy Degradation & -1.6\% & -4\% \\
        \bottomrule
    \end{tabular}
    \caption{Observed accuracy degradation across multi-round interactions.}
    \label{tab:accuracy_degradation}
\end{table}

\subsection{Cost Analysis}
\label{appendix:experiment_3}
We calculate the average cost per evaluation round. Table~\ref{tab:cost_comparison} shows that the cost of our framework is comparable to other multi-turn evaluation methods. For fair comparison, we normalized all costs to the scenario of using GPT-4-turbo as the evaluator based on the reported token usage and pricing~\cite{wang2023mint,yu-etal-2024-kieval}.

\begin{table}[h]
    \centering
    \small
    \begin{tabular}{l c}
        \hline
        \textbf{Method} & \textbf{Cost (USD)} \\
        \hline
        Human~\cite{wang2023mint} & 2.437 \\
        LLM-Judge (Single-turn) & 0.0103 \\
        MINT (Multi-turn)~\cite{wang2023mint} & 0.0563 \\
        KI-eval (Multi-turn)~\cite{yu-etal-2024-kieval} & 0.135 \\
        Ours & 0.0720 \\
        \hline
    \end{tabular}
    \caption{Comparison of average cost per evaluation round normalized to GPT-4-turbo pricing.}
    \label{tab:cost_comparison}
\end{table}

While multi-turn evaluations are generally more expensive than single-turn methods, they provide more realistic assessments, which single-turn approaches often fail to capture~\cite{wang2023mint}. It is noted that our method is more scalable than using human evaluators.

\clearpage

\begin{table*}[t]
\small
\centering
\renewcommand{\arraystretch}{1.2} 
\begin{tabular}{@{}lp{0.7\textwidth}@{}}
\specialrule{1.2pt}{0pt}{3pt} 
\textbf{Feedback} & \textbf{Example} \\
\specialrule{1.2pt}{3pt}{3pt} 
\quad Knowledge/Factuality & Your response effectively outlines the purpose and communication methods of the AQI and is clearly articulated. To improve further, ensure that the AQI scale range examples are accurate and align with commonly accepted values. Additionally, providing a more detailed explanation of the AQI calculation and data collection process would enhance the \textbf{depth and completeness} of your answer. \\ \midrule
\quad Reasoning  & Your initial steps show some understanding of the order of operations, but you incorrectly simplified the expression inside the brackets. \textbf{Remember to first evaluate the power, then apply the subtraction}, and follow through with the multiplication and division before finally adding the constant outside the brackets. \\ 

\specialrule{1.2pt}{3pt}{3pt} 

\end{tabular}
\caption{\textbf{Examples of Feedback posed by the \Interviewer{} (GPT-4o)}. }
\label{table:real_ex_feedbac}
\end{table*}
\begin{table*}[ht]
\small
\centering
\begin{tabular}{@{}ll|l@{}}
\toprule
                         & Original                                                                                                                                                                                                                                                                                                                                                               & Modified                                                                                                                                                                                                                                                                                                             \\ \midrule
MATH                     & \multicolumn{1}{c|}{\begin{tabular}[c]{@{}c@{}}If a recipe for a two-pound cake requires \\ \textcolor{blue}{1.5} cups of flour, how many cups are \\ needed for 2 five-pound cakes?\end{tabular}}                                                                                                                                                                                       & \multicolumn{1}{c}{\begin{tabular}[c]{@{}c@{}}If a recipe for a two-pound cake requires \\ \textcolor{purple}{y} cups of flour, how many cups are \\ needed for 2 five-pound cakes?\end{tabular}}                                                                                                                                        \\ \midrule
\multirow{2}{*}{DepthQA} & \begin{tabular}[c]{@{}l@{}}Question\\ What are the \textcolor{blue}{properties of straight lines}\\  in geometry?\end{tabular}                                                                                                                                                                                                                                                           & \begin{tabular}[c]{@{}l@{}}Question\\ What are \textcolor{purple}{parallel and perpendicular lines}\\  in geometry?\end{tabular}                                                                                                                                                                                                      \\ \cmidrule(l){2-3} 
                         & \begin{tabular}[c]{@{}l@{}}Reference Solution\\ 1. A \textbf{{straight line}} is the shortest distance (...)\\ 8. \text{Parallel lines} are always the same \\ distance apart and never meet.(...)\\ 10. Two lines on a plane that never \\ meet are called \text{parallel lines}.\\ 11. Two lines that intersect at a right angle \\ (90 degrees) are called \text{perpendicular lines}.\end{tabular} & \begin{tabular}[c]{@{}l@{}}Reference Solution\\ 1. A {straight line} is the shortest distance (...)\\ 8. \textbf{Parallel lines} are always the same \\ distance apart and never meet.(...)\\ 10. Two lines on a plane that never \\ meet are called \textbf{parallel lines}.\\ 11. Two lines that intersect at a right angle \\ (90 degrees) are called \textbf{perpendicular lines}.\end{tabular} \\ \bottomrule
\end{tabular}
\caption{\textbf{Example of Query Modification. }Changed parts are colored \textcolor{blue}{blue} in the original and \textcolor{purple}{purple} in the modified question.}
\label{tab:preinterview-ex}
\end{table*}
\begin{table*}[ht]
\centering
\begin{tabular}{@{}lccc@{}}
\toprule
& & {\textbf{Interviewer Model}} &  \\ \cmidrule{2-4}
Interview Phase            & \textbf{GPT-4o} & \textbf{Llama-3.1-70B} & \textbf{Llama-3.1-8B} \\ \midrule
Query Modification         & 89.5\% (105)  & 84\% (100)  & 73\% (100)  \\ \midrule
Feedback Generation        & 85.8\% (162)  & 73.5\% (102) & 58.1\% (105) \\ \midrule
Follow-up Question Generation & 93.8\% (145)  & 85\% (100)  & 62\% (100)  \\ \bottomrule
\end{tabular}
\caption{Each Interviewer's Performance Across Different Phases (\%)}
\label{tab:interviewer_performance}
\end{table*}

\clearpage

\section{Prompt\footnote{Not all prompts are included here. Refer to our GitHub repository for additional prompts.}}
\label{appendix:prompt}
\subsection{Preparing Seed Questions}
\label{appendix:prompt_1}
\begin{tcolorbox}[colback=white, colframe=gray!50, title=MATH]
You are given a mathematical expression where key information must be removed, but the remaining structure and operations should not change. Only remove critical data that makes the problem unsolvable without that information.

Guidelines:\\
1. Remove only key information required for the solution (e.g., constants, values) without changing the operations and convert to unknown variables or ambiguous words. \\
2. Ensure the revised question retains the same structure and other information.
\\ 3. Avoid removing trivial or subtle details.
\\ 4. Do not modify the mathematical operations or alter the structure of the equation.
\\ 5. In the explanation, detail exactly how and where deleted information fits into the equation or text. For example, explain which unknown variable or ambiguous words denote what value.

Response format must be in JSON as shown below:

\textbf{Example 1:}
{\{Example\}}\\
\textbf{Example 2:}
{\{Example\}}\\
\textbf{Input Question:} {\{Question\}}

\textbf{Output:}
\end{tcolorbox}

\begin{tcolorbox}[colback=white, colframe=gray!50, title=DepthQA,breakable]
You are provided with a question and its solution. Your task is to create a unique question based on these.

\textbf{Instructions:}
\\1. Create one new question that is distinct from the original question:
\\- The new question should be directly answerable using the provided solution.
\\ - It should avoid focusing on overly minor or trivial details.
\\2. Ensure the new question aligns well with the given solution and maintains relevance to the original context.

\textbf{Output Format:}
{\{Format Example\}}

\textbf{Input:}\\
Question: {\{Question\}} \\
Solution: {\{Solution\}}
\end{tcolorbox}

\subsection{Interview Process}
\label{appendix:prompt_2}
\subsubsection{Feedback Generation}

The feedback generation follows two steps approach. First, the response is graded (using the Grader Prompt below), and then feedback for the \Interviewee{} is generated based on that grading (using the Feedback Generator Prompt below).
\begin{tcolorbox}[colback=white, colframe=gray!50, title=MATH Grader,breakable]
Task: You are provided with a model’s output, a reference solution, and a correct answer. Your goal is to determine the correctness of the model's output by comparing it with the reference solution and the correct answer. Note that the reference solution is just a guide—there could be other valid methods to solve the problem.

\textbf{Possible Error Types:}\\
1. \textbf{Concept}: This error type indicates the model lacks the concept used to solve the problem. In this case, the feedback should contain a question that can check the understanding of the concept. (e.g., What is the Pythagorean theorem?) \\
2. \textbf{Misinterpret}: The model misunderstood or misinterpreted the question. In this case, the feedback should include a follow-up question or clarification that helps the model reassess and correctly interpret the original question. This ensures the model understands the context and requirements before proceeding. \\
3. \textbf{Calculation}: The model made a mistake in calculation. \\
4. \textbf{N/A}: None of the above.

\textbf{Response Format:}
{\{Format Example\}}

\textbf{Input:} \\
Initial question: \{Question\} \\
Correct answer: \{answer\} \\
Reference solution: \{Solution\} \\
Model's Output: {history}
\end{tcolorbox}

\begin{tcolorbox}[colback=white, colframe=gray!50, title=MATH Feedback Generator,breakable]
Refer to the pre-generated evaluation on the given question below and generate feedback for the model.

Your response must include:
1. Feedback: Provide the model with concise, constructive guidance based on the evaluation and reference solution. \\
2. Feedback Type: Choose one of the following: \\
   - Conceptual Guidance: Focus on understanding key concepts. \\
   - Error Identification and Correction: Address specific mistakes and guide corrections. \\
   - Process and Strategy Guidance: Improve the model’s approach or strategy. \\
   - Precision and Accuracy Emphasis: Stress precision in calculations or answer format. \\
   - Encouragement and Affirmation: Motivate and reinforce correct actions.

\textbf{Guidelines:}
1. Do not reveal the solution. Guide the assistant toward understanding the correct approach. \\
2. Ensure feedback is unique, avoiding repetition. \\
3. Provide progressively more specific hints without focusing on trivial issues. \\
4. If the model does not seem to understand the question, explain what the question is.

\textbf{Response format must be in JSON as shown below:}
{\{example\}}

\textbf{Input:}\\
Question: \{Question\} \\
Reference Solution (DO NOT disclose): \{Solution\} \\
Correct Answer (DO NOT disclose): \{answer\} \\
Pre-generated Evaluation: {evaluation} \\
Previous Feedback: {model history} \\
Model Output (to evaluate): {model output}

\textbf{Output:}
\end{tcolorbox}
\begin{tcolorbox}[colback=white, colframe=gray!50, title=STEM Grader,breakable]
\textbf{Task:} \\
You are given a question, a reference solution, and the model's output, which is broken down into atomic fact units, along with their correctness and justification. The model has revised the incorrect parts of its original output, and these revisions are provided in the model's correction statement.

Your task is to update the model's original output by replacing the corresponding revised facts with those found in the model's correction statement. Ensure that updated facts are from the model's correction statement, not the reference solution. For facts that were not revised, maintain the same content and correctness as before, so that your output contains the same number of atomic facts as the model's original output.

\textbf{Output Format:}
\{Format Example\}

\textbf{Input:}\\
Question: \{question\} \\
Solution: \{solution\} \\
Previous Feedback: \{feedback\} \\
Model's original output with correctness: \{history\} \\
Model's correction statement: \{correction\} \\

\textbf{Output:}
\end{tcolorbox}
\begin{tcolorbox}[colback=white, colframe=gray!50, title=STEM Feedback Generator,breakable]
You are an expert tasked with evaluating and providing feedback on an assistant's performance. Your response should contain the following elements:\\
1. \textbf{"status"}: Indicate whether the answer is "complete" (fully correct), "partially\_correct" (some elements are correct but not all), or "incorrect" (entirely wrong). \\
2. \textbf{"feedback"}: Provide specific, constructive feedback based on the differences between the correct solution and the model's output. Your feedback should: \\
   - Highlight what aspects of the answer are correct (if any). \\
   - Identify specific areas where the model's output differs from the correct solution, without revealing the entire solution. \\
   - Offer guidance or hints that can help the model improve its answer, focusing on the concepts or steps that need correction. \\
   - If the process is incorrect but the final answer is right, explain that the reasoning needs improvement. \\
   - Encourage the model to think step-by-step and retry if necessary.

\textbf{Remember:}
- Do not give away the complete solution or tell exactly which step is incorrect. \\
- If it's not the first attempt, provide more detailed hints, such as mentioning a relevant equation or concept, without repeating previous feedback. \\
- Tailor your feedback to the specific errors or misconceptions evident in the model's output. \\
- If there is a lack in at least one aspect among completeness, redundancy, readability, or depth, consider it incomplete and provide feedback on the identified shortcomings.

\textbf{Examples of the expected format and style of feedback:}
\{Examples\}\\
\textbf{Input:}\\
Question: {question} \\
Reference Answer (DO NOT disclose this to the assistant): {answer} \\
Correct Solution (DO NOT disclose this to the assistant): {solution} \\
Previous Feedbacks: {model\_history} \\
Model Output (This is the part to be evaluated): {model\_output}

Ensure your feedback is unique and does not repeat previous feedback.

\textbf{Expert Feedback:}
\end{tcolorbox}

\subsubsection{Follow-Up Question Generation}
\textbf{MATH Fail:} \\
If the seed question is answered incorrectly, generate a clarification question based on the model's error type. The question should focus on the concept or interpretation of the seed question to help identify the source of the misunderstanding.\\
\textbf{MATH Success:} \\
If the seed question is answered correctly, generate a follow-up rationale question to assess the model's understanding of the reasoning behind its solution.\\
\textbf{DepthQA:} 
Regardless of whether the seed question is answered correctly or not, generate a follow-up question based on the reference material. The question should inquire about additional facts or details that the model did not address in its response.

\begin{tcolorbox}[colback=white, colframe=gray!50, title=MATH Fail,breakable]
\textbf{Task:} \\
You are a question generator for assessing the model. Below are the list of the answer history of an Evaluatee model that keeps getting the question wrong, its error type, the corresponding problem, and the solution. Your role is to generate a proper follow-up question based on the model's error type. 

\textbf{Instructions:}
1. Model's Error Type is \texttt{\{error$\_$type\}}. Do not disclose the error type to the model. \\
2. Based on the error type, generate appropriate feedback as described below.

\textbf{Possible Error Types:}\\
1. \textbf{Concept:} This error type indicates the model lacks the concept used to solve the problem. The feedback should contain a question to assess the understanding of the concept. (e.g., What is the Pythagorean theorem?) \\
2. \textbf{Misinterpret:} The model misunderstood or misinterpreted the question. The feedback should include a follow-up question or clarification to help the model reassess and correctly interpret the original question. This ensures the model understands the context and requirements before proceeding. \\
3. \textbf{Calculation:} The model made a mistake in calculation. Feedback should focus on identifying the part that requires recalculation. \\
4. \textbf{N/A:} None of the above. Provide a custom description.

\textbf{Response Format (JSON):}
\{Format Example\}

\textbf{Input Data:} \\
Question: \{question\} \\
Correct Answer (DO NOT disclose): \{answer\} \\
Reference Solution (DO NOT disclose): \{solution\} \\
Previous Feedback: \{Dialogue\_History\} \\
Model's Error Type: \{error\_type\} \\

\textbf{Output:}
\end{tcolorbox}

\begin{tcolorbox}[colback=white, colframe=gray!50, title=MATH Success,breakable]
\textbf{Task:} \\
You are an evaluator assessing the Evaluatee model. The Evaluatee has successfully answered the problem, and the following is the Evaluatee's solution. Your role is to determine whether the Evaluatee truly understands their solution or if they arrived at the answer through memorization without proper understanding. To do this, you should ask questions that can assess the model's understanding of its own solution.

\textbf{Guidelines:}
1. If there are errors or missing steps in the Evaluatee's solution, ask questions to clarify or correct these issues. \\
2. If there are no errors in the solution, ask questions to confirm the Evaluatee's understanding of the solution steps. \\
3. Generate one question and provide the answer in JSON format. Be sure not to ask questions that are already present in the previous history below. \\
4. Create the question solely based on the model's solution.

\textbf{Response Format (JSON):}
\{Format Example\}

\textbf{Input Data:} \\
\textbf{Initial Question:} \{initial\_question\} \\
\textbf{Correct Answer (DO NOT disclose):} \{answer\} \\
\textbf{Reference Solution (DO NOT disclose):} \{solution\} \\
\textbf{Model's Solution:} \{model\_solution\} \\
\textbf{Output:}
\end{tcolorbox}

\begin{tcolorbox}[colback=white, colframe=gray!50, title=DepthQA,breakable]
\textbf{Task:} \\
You are given three inputs: 
1. \textbf{Question:} The original question that was asked. \\
2. \textbf{Reference Answer:} A list of atomic facts with labels indicating whether each fact is supported or unsupported by a model’s output. \\
3. \textbf{Model's Output:} The model's output and the corresponding correctness for the given question.

\textbf{Instruction:} \\
1. Your task is to generate follow-up questions based on the atomic facts from the Reference Answer that were labeled as unsupported by the model's output. \\
2. The follow-up questions should be designed to test whether the model understands the unsupported facts. \\
3. Never include any facts from the reference solution to the question. Instead, ask questions to verify knowledge. \\
4. The questions should indirectly reference the concepts in a way that allows us to assess the model’s understanding. \\
5. Your follow-up questions should not simply ask the fact directly but should guide the model to demonstrate whether it knows the fact.

\textbf{Output Format (JSON):}
\{Format Example\}

\textbf{Input Data:} \\
Question: {question} \\
Reference Answer: {solution} \\

\textbf{Output:}
\end{tcolorbox}

\subsection{Summarizing Interview}
\label{appendix:prompt_3}
\begin{tcolorbox}[colback=white, colframe=gray!50, title=Session Summarize Prompt]
\textbf{Task:} \\
The following is a conversation between \texttt{{user}} and \texttt{{system}}. Summarize the conversation.

\textbf{Instructions:} \\
1. Instead of focusing on the specific details of the questions and answers, provide a summary that highlights:\\
   - The overall flow of the conversation.\\
   - {{system}}'s problem-solving abilities.

\textbf{Input:} \\
Session History: \texttt{\{session\_history\}}

\textbf{Output:} \\
Summary:
\end{tcolorbox}
\newpage


\begin{tcolorbox}[colback=white, colframe=gray!50, title=Summarize Prompt for Interview Report,breakable]
Summarize the following problem summaries of the system's problem-solving abilities.
In your summary, try to provide \textbf{general insights} into the capabilities of the model, such as:
\\ \- Strength and weakness of the model
\\ \- Does the model respond well to the user's follow-up questions?
\\ \- Does the model handle the user's feedback effectively?
\\ \- What types of information or tasks does the model handle well?
\\ \- What types of problems or details does the model struggle with?

Answer the following questions one by one.
Do \textbf{not} focus on specific examples, but rather offer a general overview of the model's strengths and weaknesses.

Here are the system's problem-solving abilities to summarize. Don't generate too long:
\texttt{\{chunk\_dict\}}
\end{tcolorbox}

\section{Evaluating LLMs with
LLM-as-an-Interviewer}
\label{appendix:simulate}
\subsection{Evaluation Result of MINT}
\label{appendix:simulate_mint}
\begin{table*}[t]
\centering
\small
\begin{tabular}{@{}l|ccccc@{}}
\toprule
\textbf{Model} & \textbf{Judge} & \textbf{Acc\_seed(1)} & \textbf{Acc\_seed(2)} & \textbf{Acc\_seed(3)} & $\Delta$ \\
\midrule
\textbf{GPT-4o} & 0.753 & 0.753 & 0.832 & 0.858 & 0.105 \\
\textbf{GPT-3.5-turbo} & 0.467 & 0.467 & 0.644 & 0.717 & 0.250 \\
\textbf{Llama-3.1-70b} & 0.648 & 0.648 & 0.759 & 0.794 & 0.146 \\
\textbf{Llama-3.1-8b} & 0.537 & 0.537 & 0.673 & 0.733 & 0.196 \\
\textbf{DeepSeek-math} & 0.155 & 0.155 & 0.389 & 0.509 & 0.354 \\
\textbf{Qwen-math} & 0.215 & 0.215 & 0.304 & 0.342 & 0.127 \\
\bottomrule
\end{tabular}
\caption{\textbf{Performance of different models using LLM-as-an-Interviewer Framework in MINT reasoning dataset.} $\Delta$ represents the difference between Acc\_seed(3) and Acc\_seed(1).}
\label{tab:mint_result}
\end{table*}

\begin{table*}[t]
\centering
\small
\begin{tabular}{@{}l|cccc|c@{}}
\toprule
\textbf{Model} & \textbf{k = 1} & \textbf{k = 2} & \textbf{k = 3} & \textbf{k = 5} & $\Delta$ \\
\midrule
\textbf{GPT-3.5-turbo} & 0.0316 & 0.117 & 0.323 & 0.535 & 0.2914 \\
\textbf{Llama-3.1-70b} & 0.370 & 0.411 & 0.484 & 0.658 & 0.114 \\
\textbf{Llama-3.1-8b} & 0.282 & 0.358 & 0.364 & 0.377 & 0.082 \\
\bottomrule
\end{tabular}
\caption{\textbf{Performance of different models using MINT Framework in MINT reasoning dataset.} K denotes the number of interactions. $\Delta$ represents the difference between k = 3 and k = 1.}
\label{tab:mint-paper}
\end{table*}

As highlighted in the Related Work section, prior studies of multi-turn dynamic evaluation are designed with distinct objectives and methodologies, making direct comparisons difficult. Among the available benchmarks, we select \textbf{MINT}~\cite{wang2023mint}, which evaluates \textit{Multi-turn Interaction with Tool usage}, as it aligns closely with our focus. Specifically, we choose the \textbf{Reasoning} subset of MINT, which excludes external tools or code execution and consists of \textbf{316 samples} sourced from \textit{GSM8K, MATH, HotpotQA, TheoremQA, and MMLU}.

For this evaluation, we leverage our \textbf{LLM-as-an-Interviewer framework}, implemented as a \textbf{PyPI module}, showcasing its applicability beyond datasets like \textit{MATH} and \textit{DepthQA}.

\subsubsection{Evaluation}
We assess six models and compare three models—Llama-\{8B/70B\} and GPT-3.5-Turbo—that have prior results reported in MINT. To maintain consistency in measuring accuracy \textbf{without feedback or interaction}—comparing \textbf{Acc\_seed(1) from our framework with k=1 from MINT}—we do not apply any query modifications.

Table~\ref{tab:mint_result} presents the performance assessed by the LLM-as-an-Interviewer framework, while Table~\ref{tab:mint-paper} shows the results from the MINT framework. Discrepancies between \textbf{Acc\_seed(1) and k=1} may arise due to differences in \textit{tool usage and prompt structures}. MINT employs \textit{longer initial prompts} that explicitly guide tool usage and define output formats, potentially influencing accuracy at \textit{k=1}.

\subsection{Evaluation Result of MT-bench}
\label{appendix:simulate_mtbench}
We compare performance on MT-Bench~\cite{chiang2024chatbotarenaopenplatform}, a 2-turn static benchmark with 80 samples across various tasks (e.g., writing, reasoning, math, coding, roleplay). 
For the baseline, model responses are evaluated sequentially using turn = 1 and turn = 2 queries from the dataset. In the \Interviewer{} setting, turn = 1 queries serve as seed questions without modification, followed by follow-up questions posed by the \Interviewer{} (turn = 2). 
The scores range from 1 to 10 (refer to the method in~\cite{chiang2024chatbotarenaopenplatform}).

\begin{table}[h]
\centering
\small

\begin{tabular}{lcccc}
\toprule
\textbf{Model} & \multicolumn{2}{c}{\textbf{Baseline}} & \multicolumn{2}{c}{\textbf{\Interviewer{}}} \\
 & 1 & 2 & 1 & 2 \\
\midrule
GPT4o & 9.08 & 8.30 & 9.03 & 8.52 \\
GPT-3.5-turbo  & 8.40 & 7.54 & 8.38 & 8.18 \\
Llama-3.1-70B  & 8.70 & 8.08 & 8.70 & 8.36 \\
Llama-3.1-8B   & 8.52 & 7.58 & 8.48 & 7.75 \\
Deepseek-Math-7B  & 5.68 & 5.56 & 5.51 & 5.71 \\
Qwen-Math-7B   & 4.91 & 2.73 & 4.6 & 4.9 \\
\bottomrule
\end{tabular}
\caption{Performance comparison of models on MT-Bench and the \Interviewer{} setting.}
\label{table:mtbench_results}
\end{table}

\subsection{Follow-Up Question Example of DepthQA}
\label{appendix:simulate_2}
In the following example, the follow-up question targets the n-th roots mentioned in the reference solution but missing from the model’s previous response. This type of follow-up question is central to DepthQA. It partially address a key limitation of traditional factuality scoring, which only evaluates the facts the model provides (i.e., precision). By probing further with additional questions, we can assess the model’s broader factuality, capturing missing or incomplete information.
\newpage
\begin{tcolorbox}[colback=white, colframe=gray!50!black, title=DepthQA Follow-up Question Example,breakable]
\noindent \textbf{User:} What is the significance of the roots of unity in mathematics and physics? \\
\textbf{System:} The roots of unity are important in mathematics and physics because they have connections to many different areas of study. In mathematics, they are used in algebra, number theory, and complex analysis. In physics, they appear in wave functions, quantum mechanics, and signal processing. The roots of unity also have applications in computer science, cryptography, and engineering. Overall, they play a key role in understanding the behavior of periodic functions and symmetries in various mathematical and physical systems.

\vspace{10pt}
\texttt{\# Follow-up Question} \\
\noindent \textbf{{User:}} Can you provide an example of determining the \textbf{6th roots of unity} and specifying their arguments in radians? \\

\noindent \hrulefill
\newpage
\noindent \textbf{Reference Solution}

The roots of unity refer to the complex solutions of the equation \( z^n = 1 \), where \( n \) is a positive integer. These roots are regularly spaced around the circumference of the unit circle in the complex plane and hence, are of significant importance in many areas of mathematics and physics.

The term "unity" signifies the number 1, and "roots" refer to the solutions of an equation. So, the roots of unity are essentially the solutions to the equation that equal 1 when raised to the power \( n \).

\textbf{Each \( n \)-th root of unity is given by the formula:}
\(
z = \cos\left(\frac{2\pi k}{n}\right) + i \sin\left(\frac{2\pi k}{n}\right)
\)

for \( k = 0, 1, \dots, n - 1 \). Here, \( \cos \) and \( \sin \) are the standard trigonometric functions, \( i \) refers to the imaginary unit, and \( \pi \) is the mathematical constant Pi (approximately equal to 3.14).

Let's say if you wanted to determine the 4th roots of unity, (omitted)

\vspace{10pt}
\end{tcolorbox}
\clearpage
\clearpage
\section{Full Interview Log}
\label{appendix:full_log}
\begin{tcolorbox}[colback=white, colframe=yellow!50!yellow, title=MATH Full Log,fonttitle=\color{black},breakable]

\textbf{Interviewee : Llama-3.1-70b, Interviewer : GPT-4o} \\

\textbf{User:} Question: The number \( \sqrt{x \sqrt{6} + y \sqrt{10} + z \sqrt{15} + w} \) \\
can be written as \( a \sqrt{2} + b \sqrt{3} + c \sqrt{5} \), where \( a \), \( b \), and \( c \) \\
are positive integers. Find \( a \cdot b \cdot c \). \\

\textbf{} The given expression can be written as \( a \sqrt{2} + b \sqrt{3} + c \sqrt{5} \), \\
where \( a \), \( b \), and \( c \) are positive integers. \\
We are given the expression: \\
\( \sqrt{x \sqrt{6} + y \sqrt{10} + z \sqrt{15} + w} \) \\
Squaring the expression to get rid of the outer square root: \\
\( (x \sqrt{6} + y \sqrt{10} + z \sqrt{15} + w) = (a \sqrt{2} + b \sqrt{3} + c \sqrt{5})^2 \) \\
Expanding the right side: \\
\( (x \sqrt{6} + y \sqrt{10} + z \sqrt{15} + w) = 2a^2 + 3b^2 + 5c^2 + 2ab \sqrt{6} + 2ac \sqrt{10} + 2bc \sqrt{15} + 2abc \sqrt{30} \) \\
Comparing the coefficients of \( \sqrt{6} \), \( \sqrt{10} \), and \( \sqrt{15} \): \\
\( x = 2ab \), \( y = 2ac \), \( z = 2bc \) \\
Also, the constant term \( w \) is: \\
\( w = 2a^2 + 3b^2 + 5c^2 \) \\
We know that \( a \), \( b \), and \( c \) are positive integers. \\
From the coefficients of \( \sqrt{6} \), \( \sqrt{10} \), and \( \sqrt{15} \): \\
\( x = 2ab \), \( y = 2ac \), \( z = 2bc \) \\
We can see that \( x \), \( y \), and \( z \) can be expressed as multiples of 2. \\
Also, the constant term \( w \) is: \\
\( w = 2a^2 + 3b^2 + 5c^2 \) \\
Since \( x \), \( y \), and \( z \) are multiples of 2, \( w \) must be a multiple of 2 as well. \\
The least possible values for \( a \), \( b \), and \( c \) that satisfy the conditions are: \\
\( a = 1 \), \( b = 1 \), \( c = 1 \) \\
Substituting these values in: \\
\( x = 2ab = 2(1)(1) = 2 \) \\
\( y = 2ac = 2(1)(1) = 2 \) \\
\( z = 2bc = 2(1)(1) = 2 \) \\
\( w = 2a^2 + 3b^2 + 5c^2 = 2(1)^2 + 3(1)^2 + 5(1)^2 = 10 \) \\
However, \( w = 10 \) is not a multiple of 2 that satisfies the conditions. \\

\textbf{User:} You got a correct answer! Good Job. \\

\textbf{} I'm glad I was able to solve the problem correctly in the end. \\ It was a bit of a journey, \\but using the exact values of 104, 468, 144, and 2006 for \( x \), \( y \), \( z \), and \( w \) respectively, \\
helped to simplify the problem and arrive at the correct solution. \\
If you have any more problems you'd like help with, feel free to ask! \\

\textbf{User:} How do you determine the values of 'a', 'b', and 'c' \\
from the equations \( ab = 52 \), \( ac = 234 \), and \( bc = 72 \)? \\

\textbf{} To determine the values of 'a', 'b', and 'c' from the equations \\ \( ab = 52 \), \( ac = 234 \), and \( bc = 72 \), we can use the following steps: \\
1. Divide the equation \( ac = 234 \) by the equation \( ab = 52 \): \\
   \( \frac{ac}{ab} = \frac{234}{52} \) \\
   \( \frac{c}{b} = \frac{9}{4} \) \\
2. Multiply both sides by \( b \) to get: \\
   \( c = \frac{9}{4}b \) \\
3. Now, substitute this expression for \( c \) into the equation \( bc = 72 \): \\
   \( b \cdot \left( \frac{9}{4}b \right) = 72 \) \\
   \( \frac{9}{4}b^2 = 72 \) \\
4. Multiply both sides by \( \frac{4}{9} \) to get: \\
   \( b^2 = 72 \cdot \frac{4}{9} \) \\
   \( b^2 = 32 \) \\
5. Take the square root of both sides: \\
   \( b = \sqrt{32} \) \\
   \( b = 4 \) (since \( b \) is a positive integer) \\
6. Now that we have the value of \( b \), we can find the value of \( a \) using the equation \( ab = 52 \): \\
   \( a = \frac{52}{b} \) \\
   \( a = \frac{52}{4} \) \\
   \( a = 13 \) \\
7. Next, find the value of \( c \) using the equation \( ac = 234 \): \\
   \( c = \frac{234}{a} \) \\
   \( c = \frac{234}{13} \) \\
   \( c = 18 \) \\
These values satisfy the original equations \( ab = 52 \), \( ac = 234 \), and \( bc = 72 \). \\

\textbf{User:} Your explanation is generally correct, but there are some inconsistencies in your steps. \\
Specifically, your calculation of \( b \) and the adjustment for \( c \) requires clarification. 
\\You initially found \( b = 4 \), which is correct, but then multiplied \( b \) by 2 without clear reasoning. \\
Finally, you identified the correct values, but the intermediate explanations
\\ need to be streamlined for consistency. Could you revisit \\and clarify these steps to arrive at a consistent conclusion? \\

\textbf{} You are correct, my initial explanation had some inconsistencies. \\
Here's a revised explanation for determining the values of 'a', 'b', and 'c' \\ from the equations \( ab = 52 \), \( ac = 234 \), and \( bc = 72 \): \\
1. Divide the equation \( ac = 234 \) by the equation \( ab = 52 \): \\
   \( \frac{ac}{ab} = \frac{234}{52} \) \\
   \( \frac{c}{b} = \frac{9}{4} \) \\
2. Multiply both sides by \( b \) to get: \\
   \( c = \frac{9}{4}b \) \\
3. Now, substitute this expression for \( c \) into the equation \( bc = 72 \): \\
   \( b \cdot \left( \frac{9}{4}b \right) = 72 \) \\
   \( \frac{9}{4}b^2 = 72 \) \\
4. Multiply both sides by \( \frac{4}{9} \) to get: \\
   \( b^2 = 72 \cdot \frac{4}{9} \) \\
   \( b^2 = 32 \) \\
5. Take the square root of both sides: \\
   \( b = \sqrt{32} \) \\
   \( b = 4 \) (since \( b \) is a positive integer) \\
6. Now that we have the value of \( b \), we can find the value of \( a \) using the equation \( ab = 52 \): \\
   \( a = \frac{52}{b} \) \\
   \( a = \frac{52}{4} \) \\
   \( a = 13 \) \\
7. Next, find the value of \( c \) using the equation \( ac = 234 \): \\
   \( c = \frac{234}{a} \) \\
   \( c = \frac{234}{13} \) \\
   \( c = 18 \) \\
This way, we arrive at the correct values for \( a \), \( b \), and \( c \) in a consistent and straightforward manner: \\
   \( a = 13 \), \( b = 4 \), \( c = 18 \). \\
Thank you for pointing out the inconsistencies in my initial explanation.\\
I hope this revised explanation helps clarify the steps. \\

\end{tcolorbox}
\newpage
\begin{tcolorbox}[colback=white, colframe=green!40!gray, title=DepthQA Full Log,fonttitle=\color{white},breakable]
\textbf{Interviewee : Llama-3.1-70b, Interviewer : GPT-4o} \\ \\
\textbf{User}: Question: What are some scenarios where the forward or backward difference approximations are preferred over the central difference approximation? \\

\textbf{System}: There are several scenarios where the forward or backward difference approximations are preferred over the central difference approximation:

1. **Boundary conditions**: When approximating the derivative at a boundary point, the central difference approximation is not feasible because it requires points on both sides of the boundary. In such cases, the forward or backward difference approximation is used, depending on whether the boundary point is at the beginning or end of the interval.

2. **Non-uniform grids**: When the grid points are not uniformly spaced, the central difference approximation may not be the most accurate choice. In such cases, the forward or backward difference approximation can be used, taking into account the non-uniform spacing of the grid points.

3. **Noise or irregular data**: When the data is noisy or irregular, the central difference approximation may amplify the noise or irregularities. In such cases, the forward or backward difference approximation can be used to reduce the effect of noise or irregularities.

4. **Stability and convergence**: In some numerical methods, such as finite difference methods for solving partial differential equations, the forward or backward difference approximation may be preferred over the central difference approximation for stability and convergence reasons.

5. **Computational efficiency**: In some cases, the forward or backward difference approximation may be computationally more efficient than the central difference approximation, especially when the number of grid points is large.

6. **Advection-dominated problems**: In problems where advection dominates diffusion, the backward difference approximation may be preferred over the central difference approximation to reduce numerical diffusion.

7. **Time-dependent problems**: In time-dependent problems, the backward difference approximation may be preferred over the central difference approximation to ensure stability and convergence.

In summary, the choice of difference approximation depends on the specific problem, the nature of the data, and the requirements of the numerical method.
\\
\textbf{User}: Your explanation covers several important scenarios where forward or backward difference approximations might be preferred, but some areas need more detailed context and justification. For instance, could you explain when grid points are not uniformly spaced and why central difference may not be accurate there? Also, clarify how the central difference method might amplify noise or irregularities compared to other methods. Additionally, elaborating on why the backward difference might be used in advection-dominated problems and the suitability of backward difference for time-dependent problems would strengthen your response. Overall, consider diving deeper into these points to improve the completeness and depth of your answer. \\

\textbf{System}: I'd be happy to provide more detailed explanations for the scenarios where the forward or backward difference approximations are preferred over the central difference approximation.

**Non-uniform grids:**

When the grid points are not uniformly spaced, the central difference approximation may not be the most accurate choice. This is because the central difference formula assumes that the grid points are evenly spaced, and it uses the average of the function values at the neighboring points to approximate the derivative. However, when the grid points are not uniformly spaced, this assumption is no longer valid.

For example, consider a scenario where the grid points are spaced more closely near the boundary of the domain than in the interior. In this case, the central difference formula would use the same weighting for the function values at the neighboring points, even though the points near the boundary are more closely spaced. This can lead to inaccurate results, especially if the function is rapidly changing near the boundary.

In contrast, the forward or backward difference approximations can be adapted to handle non-uniform grids by taking into account the spacing between the grid points. For example, the forward difference formula can be modified to use the distance between the current point and the next point, rather than assuming a uniform spacing.

**Noise or irregular data:**

When the data is noisy or irregular, the central difference approximation may amplify the noise or irregularities. This is because the central difference formula uses the difference between the function values at the neighboring points, which can accentuate any noise or irregularities in the data.

For example, consider a scenario where the data has a high-frequency noise component. The central difference formula would use the difference between the function values at the neighboring points, which would amplify the noise component. In contrast, the forward or backward difference approximations would use the difference between the current point and the next or previous point, respectively, which can help to reduce the effect of the noise.

Additionally, the central difference formula can also be sensitive to outliers or irregularities in the data. For example, if there is a single outlier in the data, the central difference formula would use this outlier in the calculation of the derivative, which can lead to inaccurate results. In contrast, the forward or backward difference approximations can be more robust to outliers, as they only use the current point and the next or previous point, respectively.

**Advection-dominated problems:**

In problems where advection dominates diffusion, the backward difference approximation may be preferred over the central difference approximation to reduce numerical diffusion. Advection-dominated problems are characterized by a strong convective flow, which can lead to numerical diffusion when using the central difference approximation.

Numerical diffusion is a phenomenon where the numerical solution of the problem becomes smeared or diffused, even though the underlying physics of the problem does not involve diffusion. This can happen when the central difference approximation is used, as it can introduce artificial diffusion into the solution.

The backward difference approximation, on the other hand, can help to reduce numerical diffusion by using the upwind direction to calculate the derivative. The upwind direction is the direction in which the flow is moving, and using this direction can help to reduce the artificial diffusion introduced by the numerical method.

For example, consider a problem where a pollutant is being transported by a fluid flow. The central difference approximation would use the average of the pollutant concentrations at the neighboring points to calculate the derivative, which can lead to numerical diffusion. In contrast, the backward difference approximation would use the pollutant concentration at the upwind point to calculate the derivative, which can help to reduce numerical diffusion.

**Time-dependent problems:**

In time-dependent problems, the backward difference approximation may be preferred over the central difference approximation to ensure stability and convergence. Time-dependent problems involve solving a partial differential equation that changes over time, and the numerical method must be able to capture the time-dependent behavior of the solution.

The backward difference approximation is often preferred in time-dependent problems because it is unconditionally stable, meaning that it will converge to the correct solution regardless of the time step size. The central difference approximation, on the other hand, can be conditionally stable, meaning that it will only converge to the correct solution if the time step size is small enough.

For example, consider a problem where a wave is propagating through a medium over time. The central difference approximation would use the average of the wave amplitude at the neighboring points to calculate the derivative, which can lead to instability if the time step size is too large. In contrast, the backward difference approximation would use the wave amplitude at the previous time step to calculate the derivative, which can help to ensure stability and convergence.

In summary, the choice of difference approximation depends on the specific problem, the nature of the data, and the requirements of the numerical method. The forward or backward difference approximations can be preferred over the central difference approximation in scenarios such as non-uniform grids, noisy or irregular data, advection-dominated problems, and time-dependent problems." \\ 

\textbf{User}: Your output provides a comprehensive and detailed explanation of various scenarios where forward or backward difference approximations may be preferred over central difference. However, consider if the preference for specific difference approximations might depend on factors beyond stability and convergence, and reflect on whether computational efficiency is a significant deciding factor. Additionally, some redundancy in the discussion of non-uniform grids and noisy data could be consolidated to improve conciseness. Overall, the depth and readability of your explanation are strong, but ensure all points are necessary to the question. \\

\textbf{System}: I'd be happy to refine my explanation to address your points and provide a more concise response.

**Non-uniform grids and noisy data:**

In scenarios where the grid points are not uniformly spaced or the data is noisy or irregular, the forward or backward difference approximations may be preferred over the central difference approximation. The central difference formula assumes uniform spacing and can be sensitive to noise or irregularities, which can lead to inaccurate results. In contrast, the forward or backward difference approximations can be adapted to handle non-uniform grids and reduce the effect of noise or irregularities.

For example, in problems with high-frequency noise or outliers, the forward or backward difference approximations can provide a more robust estimate of the derivative by using the difference between the current point and the next or previous point, respectively. Similarly, in non-uniform grids, the forward or backward difference approximations can take into account the varying spacing between grid points to provide a more accurate estimate of the derivative.

**Advection-dominated problems:**

In problems where advection dominates diffusion, the backward difference approximation may be preferred over the central difference approximation to reduce numerical diffusion. The central difference approximation can introduce artificial diffusion into the solution, which can be mitigated by using the backward difference approximation and the upwind direction to calculate the derivative.

**Time-dependent problems:**

In time-dependent problems, the backward difference approximation may be preferred over the central difference approximation to ensure stability and convergence. The backward difference approximation is unconditionally stable, whereas the central difference approximation can be conditionally stable, requiring a small enough time step size to converge.

**Computational efficiency:**

While computational efficiency can be a consideration in choosing a difference approximation, it is often secondary to stability and convergence. In some cases, the forward or backward difference approximations may be computationally more efficient than the central difference approximation, especially when the number of grid points is large. However, this advantage may be outweighed by the need for stability and convergence in many applications.

**Additional considerations:**

Beyond stability and convergence, the choice of difference approximation may depend on factors such as:

* **Boundary conditions:** The forward or backward difference approximations may be preferred at boundary points, where the central difference approximation is not feasible.
* **Problem physics:** The choice of difference approximation may depend on the specific physics of the problem, such as the dominance of advection or diffusion.
* **Numerical method:** The choice of difference approximation may be influenced by the overall numerical method being used, such as finite difference or finite element methods.

In summary, the choice of difference approximation depends on a range of factors, including stability, convergence, computational efficiency, and problem-specific considerations. While the central difference approximation is often a good choice, the forward or backward difference approximations may be preferred in scenarios with non-uniform grids, noisy data, advection-dominated problems, or time-dependent problems.\\

\textbf{User}: You got a correct explanation! Good Job.\\

\textbf{User}: In the context of boundary conditions, how would you choose between forward and backward difference approximations at the start and end of the domain? \\

\textbf{System}: When choosing between forward and backward difference approximations at the start and end of the domain, it ultimately depends on the specific boundary conditions and the direction of information flow. Here's a general guideline:

**Start of the domain:**

* If the boundary condition at the start of the domain is a **Dirichlet boundary condition** (i.e., the value of the function is specified at the boundary), a **forward difference approximation** is often used. This is because the forward difference approximation looks ahead to the next point, which is inside the domain, to estimate the derivative. This approach is consistent with the direction of information flow, where the boundary condition is imposed from the outside and influences the behavior of the solution inside the domain.
* If the boundary condition at the start of the domain is a **Neumann boundary condition** (i.e., the derivative of the function is specified at the boundary), a **backward difference approximation** is not suitable, as it would look backwards outside the domain. In this case, a forward difference approximation or a ghost point method (where a fictional point is introduced outside the domain to apply the Neumann boundary condition) can be used.

**End of the domain:**

* If the boundary condition at the end of the domain is a **Dirichlet boundary condition**, a **backward difference approximation** is often used. This is because the backward difference approximation looks backwards to the previous point, which is inside the domain, to estimate the derivative. This approach is consistent with the direction of information flow, where the boundary condition is imposed from the outside and influences the behavior of the solution inside the domain.
* If the boundary condition at the end of the domain is a **Neumann boundary condition**, a **forward difference approximation** is not suitable, as it would look forward outside the domain. In this case, a backward difference approximation or a ghost point method can be used.

**Additional considerations:**

* When using forward or backward difference approximations at the boundaries, it's essential to ensure that the chosen method is consistent with the overall numerical method being used (e.g., finite difference or finite element methods).
* In some cases, a **ghost point method** or a **one-sided finite difference** can be used to apply boundary conditions. These methods introduce fictional points outside the domain to apply the boundary conditions, which can be useful when dealing with complex boundary conditions or non-uniform grids.

In summary, the choice between forward and backward difference approximations at the start and end of the domain depends on the specific boundary conditions and the direction of information flow. A forward difference approximation is often used at the start of the domain with Dirichlet boundary conditions, while a backward difference approximation is often used at the end of the domain with Dirichlet boundary conditions."
\end{tcolorbox}

\clearpage
\section{Mitigating Contamination Issue with LLM-as-an-Interviewer}
\label{appendix:contam}

\subsection{Training Details and Configurations}
\label{appendix:contam_1}
For training, we utilized four 80GB NVIDIA A100 GPUs. We trained Zephyr models using the implementation from the Alignment Handbook repository.
\footnote{\href{https://github.com/huggingface/alignment-handbook}%
{Alignment Handbook Repository}}
We trained OLMoE models using the instructions provided on its GitHub repository.
\footnote{\href{https://github.com/allenai/OLMoE}%
{OLMoE GitHub Repository}}

Training hyperparameters are detailed in Table \ref{tab:zephyr_7b_hyperparameter} and \ref{tab:olmoe_7b_hyperparameter}. 
\subsection{Experimental Results in DepthQA}
\begin{table}[ht]
\centering
\small
\begin{tabular}{@{}ll|cccc@{}}
\toprule
\multicolumn{2}{l|}{}                                            & \multicolumn{2}{c|}{\textbf{OLMoE}}                                                                   & \multicolumn{2}{c}{\textbf{Zephyr}}                                                               \\ \cmidrule(l){3-6} 
\textit{Model.}                          & \multicolumn{1}{c|}{} & \multicolumn{1}{c|}{Judge.}              & \multicolumn{1}{c|}{Interv.}                                   & \multicolumn{1}{c|}{Judge.}              & Interv.                                    \\ \midrule
\textit{\textbf{Uncontam.}}              &                       &                                         & \multicolumn{1}{c|}{}                                       &                                         &                                        \\
\hspace{3mm}$\text{Train set}_\text{ID} $           & {[}1{]}               & \cellcolor[HTML]{FFFFFF}0.70            & \multicolumn{1}{c|}{\cellcolor[HTML]{FFFFFF}0.71}               & \cellcolor[HTML]{DFF2E9}0.84            & \cellcolor[HTML]{FCFEFD}0.80               \\ \midrule
\textbf{Avg.}                              &                       & 0.70            & 0.71               & 0.84               & 0.80            \\ \midrule

\textit{\textbf{Contam.}}                &                       &                                         & \multicolumn{1}{c|}{}                                       &                                         &                                         \\
\hspace{3mm}$\text{Test set}_\text{ID} $            & {[}4{]}               & \cellcolor[HTML]{57BB8A}1.00            & \multicolumn{1}{c|}{\cellcolor[HTML]{FCFEFD}0.80}               & \cellcolor[HTML]{C9E9D9}0.86            & \cellcolor[HTML]{D7EFE3}0.85               \\
\hspace{3mm}+$\text{Train set}_\text{ID} $          & {[}5{]}               & \cellcolor[HTML]{7DCBA5}0.95            & \multicolumn{1}{c|}{\cellcolor[HTML]{FFFFFF}0.73}               & \cellcolor[HTML]{C0E6D3}0.87            & \cellcolor[HTML]{BDE4D1}0.88               \\
\hspace{3mm}+Instruct                    & {[}6{]}               & \cellcolor[HTML]{9FD8BC}0.91            & \multicolumn{1}{c|}{\cellcolor[HTML]{FFFFFF}0.70}               & \cellcolor[HTML]{CFECDE}0.86            & \cellcolor[HTML]{E9F6F0}0.83               \\ \midrule
\textbf{Avg.}                    &                       & 0.93            & 0.76               & 0.87               & 0.85            \\ \bottomrule
\end{tabular}
\caption{\textbf{Performance of uncontaminated and contaminated models on DepthQA dataset}. "Judge" refers to the LLM-as-a-Judge setting, and "Interv" refers to the LLM-as-an-Interver setting.}
\label{tab:depthqa-contam}
\end{table}

Table~\ref{tab:depthqa-contam} shows the contamination experiment results in DepthQA.
\newpage
\begin{table}[ht]
\centering
\fontsize{8}{10}\selectfont
\begin{tabular}{c|c}
\toprule
\textbf{Base Model} & alignment-handbook/zephyr-7b-sft-full\\
\textbf{Torch dtype} &  bfloat16\\
\textbf{Epoch} & 10 \\ 
\textbf{Max Seq Length} & 4096\\
\textbf{Learning Rate} & 1e-5\\
\textbf{Train Batch Size} & 32\\
\textbf{Random Seed} & 42\\
\textbf{Training Method} & Supervised Fine-tuning\\
\bottomrule
\end{tabular}%
\caption{\footnotesize Hyperparameters used to train Zephyr 7B.}
\label{tab:zephyr_7b_hyperparameter}
\end{table}

\begin{table}[ht]
\centering
\fontsize{8}{10}\selectfont
\begin{tabular}{c|c}
\toprule
\textbf{Base Model} & allenai/OLMoE-1B-7B-0924-Instruct\\
\textbf{Torch dtype} & bfloat16\\
\textbf{Epoch} & 10\\ 
\textbf{Max Seq Length} & 4096\\
\textbf{Learning Rate} & 2e-5\\
\textbf{Train Batch Size} & 16 \\
\textbf{Training Method} & Supervised Fine-tuning\\
\bottomrule
\end{tabular}%
\caption{\footnotesize Hyperparameters used to train OLMoE-1B-7B model.}
\label{tab:olmoe_7b_hyperparameter}
\end{table}


\newpage
\section{Reliability of LLM-as-an-Interviewer}
\label{appendix:Reliability}
\subsection{Verbosity Bias}
\label{appendix:Reliability_1}

\begin{figure*}[t]
    \centering
    \begin{subfigure}[t]{0.36\textwidth}
        \centering
        \includegraphics[height=4.45cm]{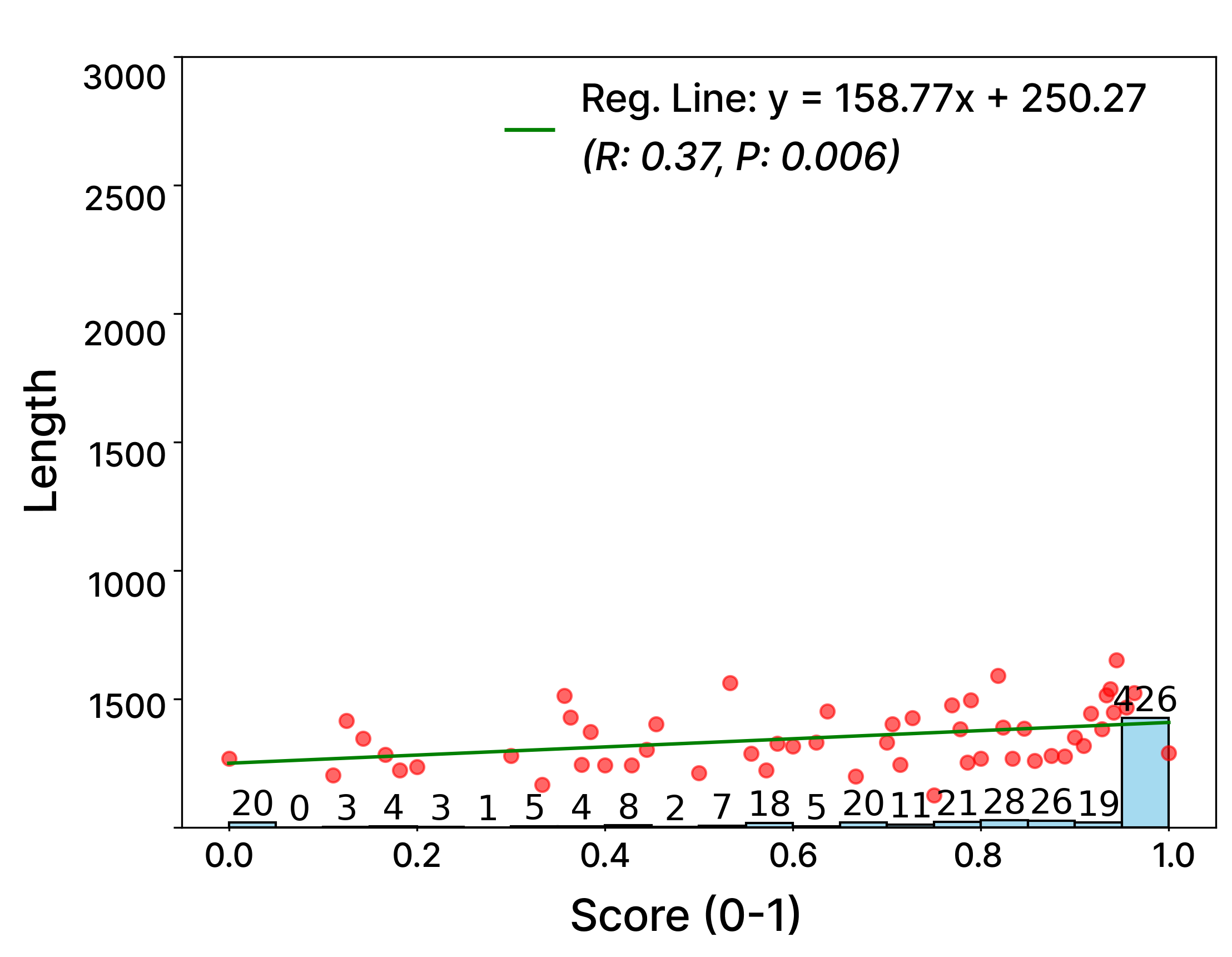}
        \caption{Interaction 1}
        \label{fig:attempt1}
    \end{subfigure}%
    \begin{subfigure}[t]{0.31\textwidth}
        \centering
        \includegraphics[height=4.45cm]{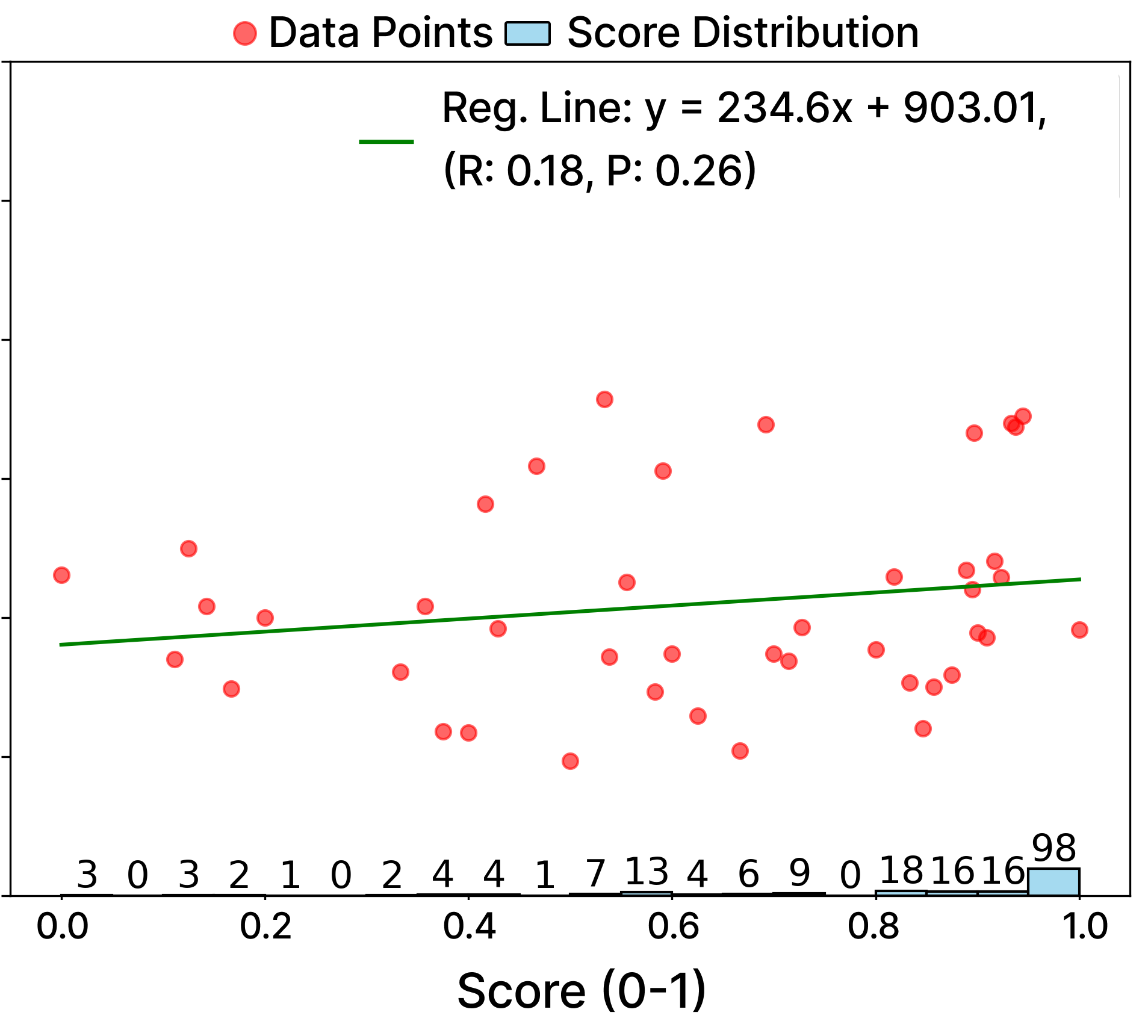}
        \caption{Interaction 2}
        \label{fig:attempt2}
    \end{subfigure}%
    \begin{subfigure}[t]{0.31\textwidth}
        \centering
        \includegraphics[height=4.45cm]{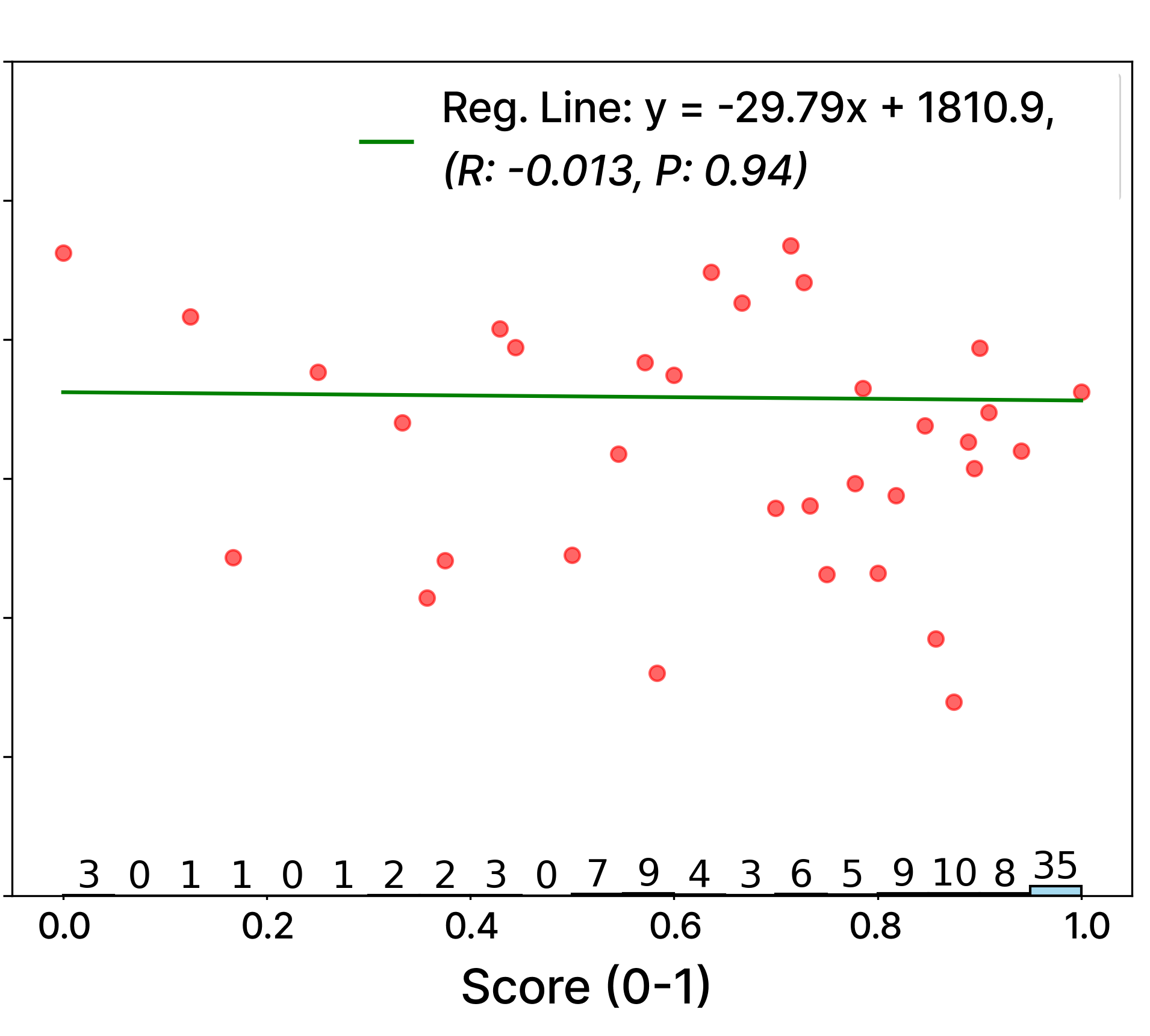}
        \caption{Interaction 3}
        \label{fig:attempt3}
    \end{subfigure}
      \caption{\textbf{Distribution of Response Lengths (measured in tokens) Across Scores in DepthQA.} The regression correlation coefficient (\textit{r}) decreases with the number of Interaction, suggesting a reduction in \textbf{Verbosity Bias} as Interactions progress. }
    \label{fig:verbosity}
\end{figure*}
Table~\ref{fig:verbosity} presents the correlation between scores and interaction lengths across different stages of the interaction.
\subsection{Self-Enhancement Bias}
\label{appendix:Reliability_2}
Existing papers primarily analyze the self-enhancement bias of LLM-as-a-Judge through relative rating—comparing the responses of two models and selecting the preferred one~\cite{chiang2024chatbotarenaopenplatform}. However, since LLM-as-an-Interviewer uses absolute rating—evaluating only a single model’s response—it is difficult to directly analyze self-enhancement bias. Therefore, we only check whether an apparently observable self-enhancement bias exists.

\begin{figure*}[ht]
    \centering
    \begin{subfigure}[t]{0.34\textwidth}
        \centering
        \includegraphics[height=4.45cm]{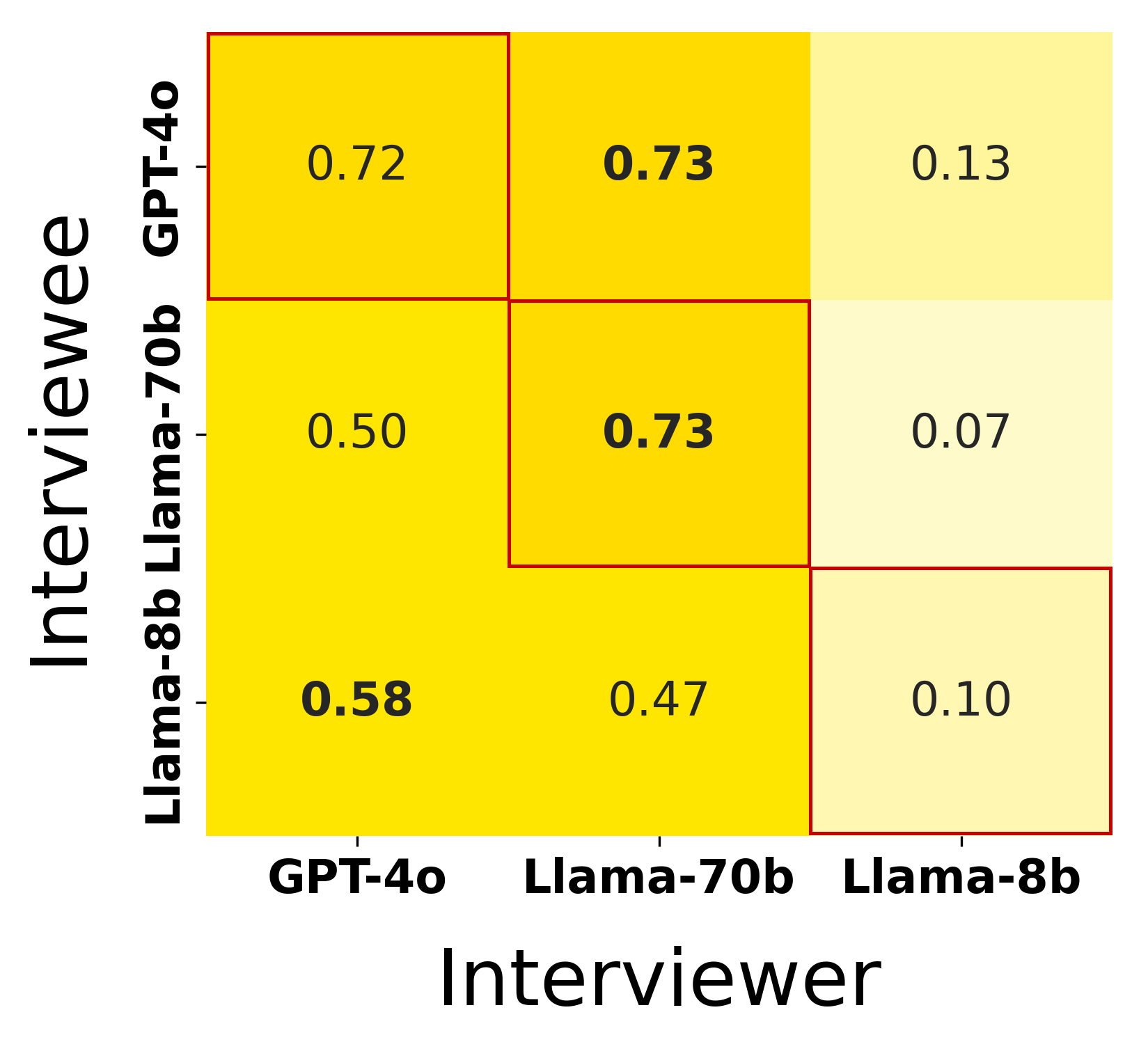}
        \caption{Interaction 1}
        \label{fig:attempt1_math}
    \end{subfigure}%
    \begin{subfigure}[t]{0.30\textwidth}
        \centering
        \includegraphics[height=4.45cm]{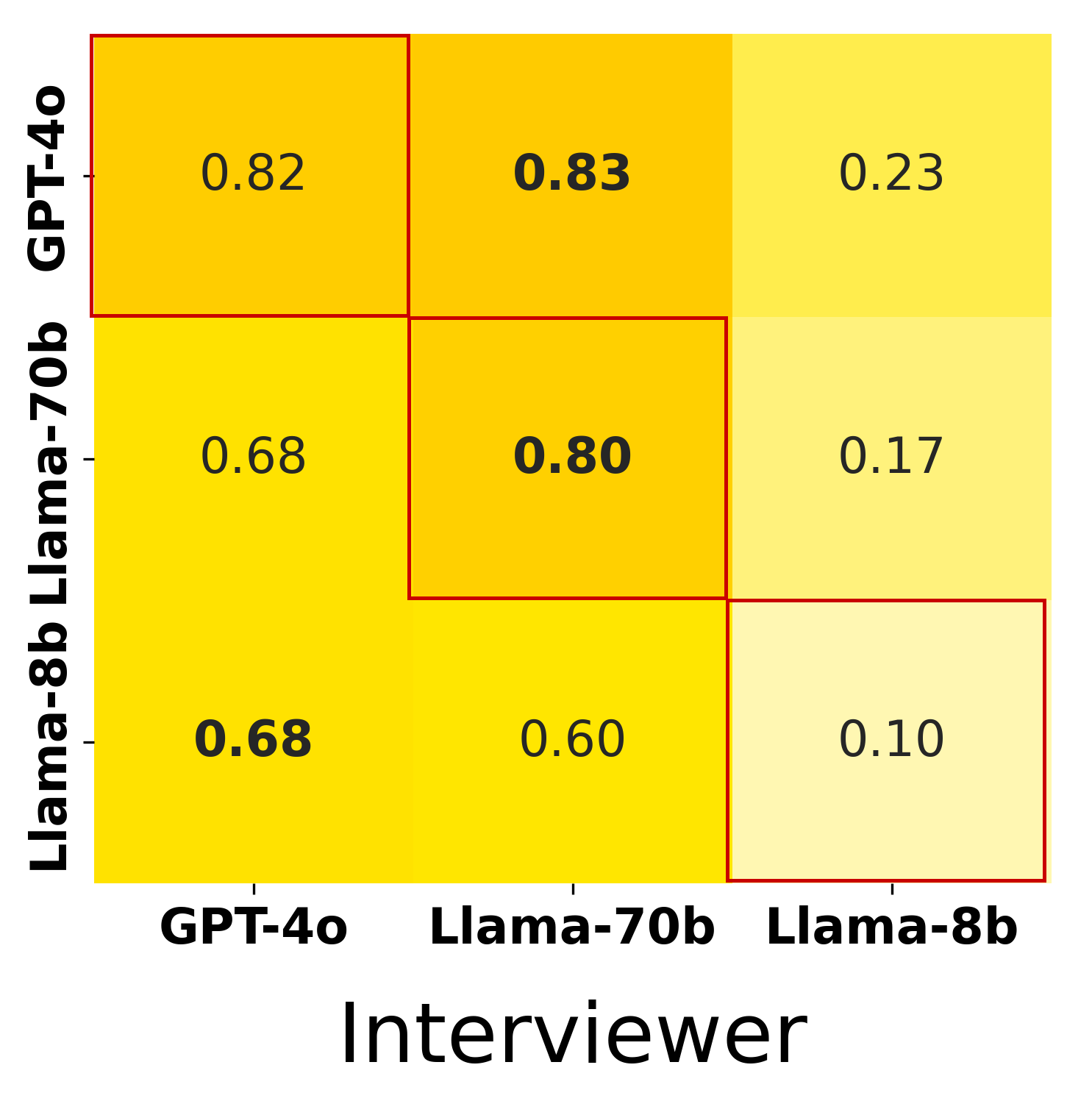}
        \caption{Interaction 2}
        \label{fig:attempt2_math}
    \end{subfigure}%
    \begin{subfigure}[t]{0.34\textwidth}
        \centering
        \includegraphics[height=4.45cm]{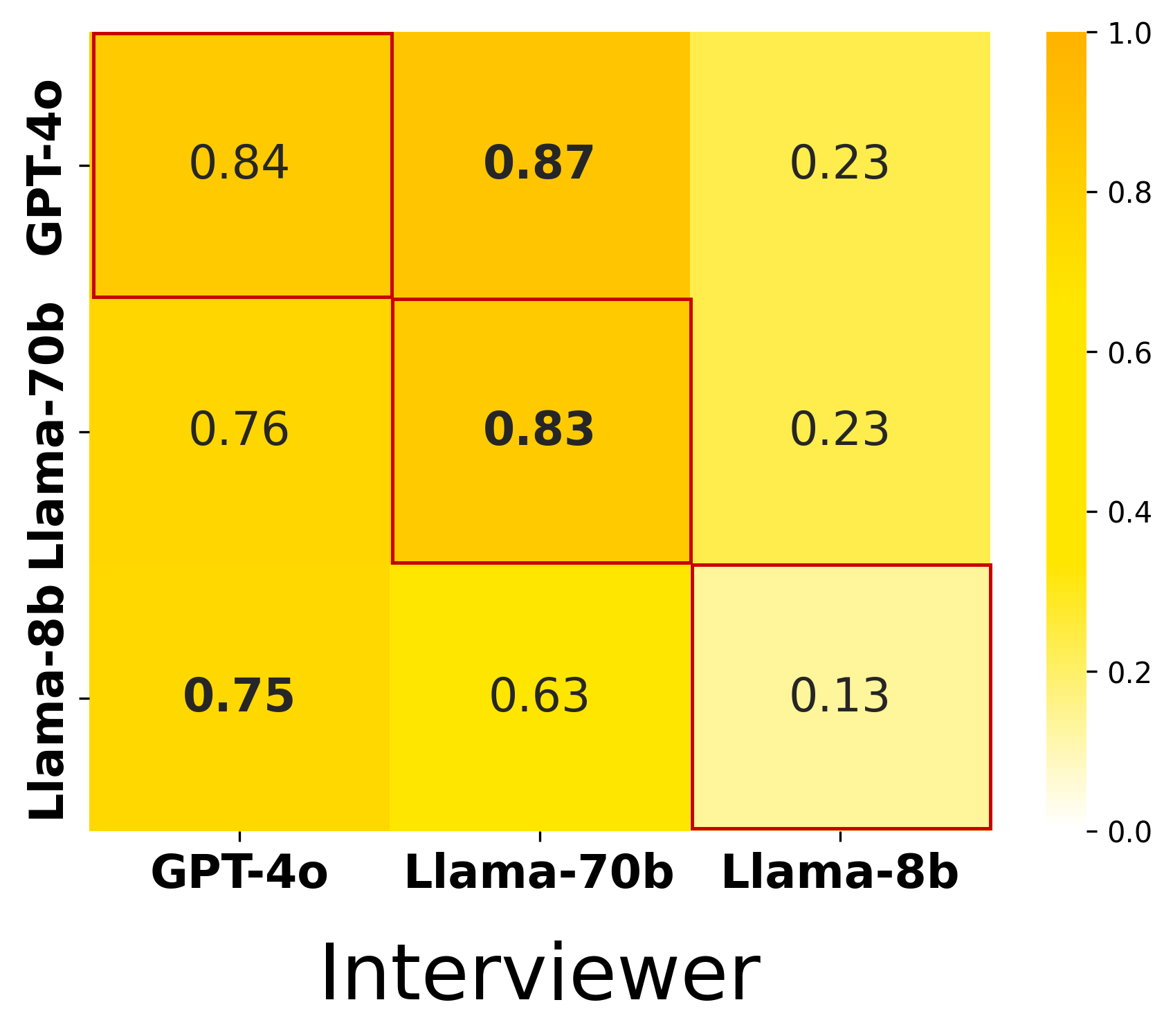}
        \caption{Interaction 3}
        \label{fig:attempt3_math}
    \end{subfigure}

    \caption{\textbf{Accuracy of Models as Interviewees by Number of Interactions in MATH}. The red-highlighted boxes indicate self-evaluated cases. Llama-70B consistently gives itself higher scores compared to other interviewers, but this does not result in a significant \textbf{Self-Enhancement Bias} that alters its ranking. No such bias is observed in the other models.}
    \label{fig:self_bias}
\end{figure*}

\begin{figure*}[ht]
    \centering
    \begin{subfigure}[ht]{0.34\textwidth}
        \centering
        \includegraphics[height=4.45cm]{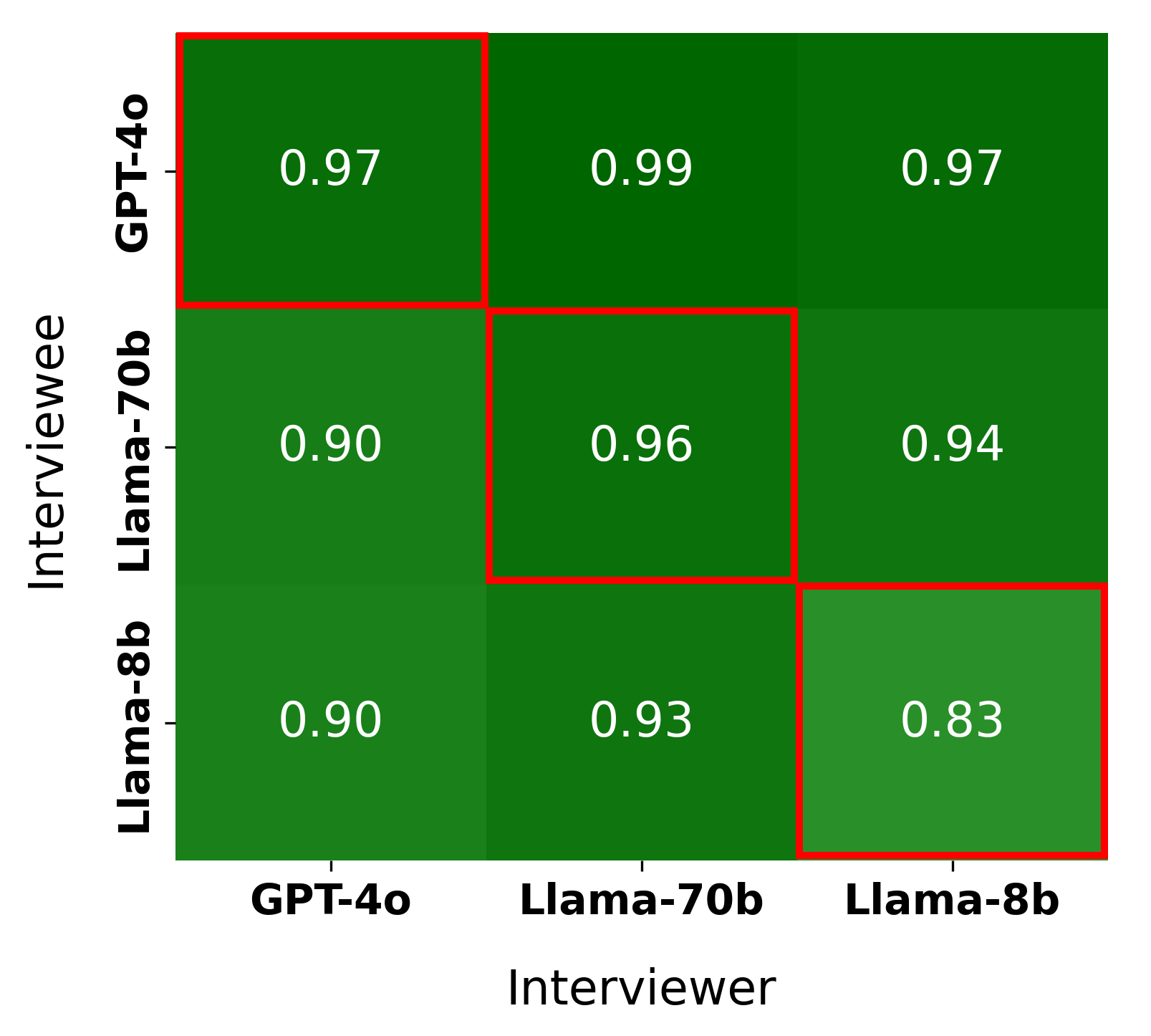}
        \caption{Interaction 1}
        \label{fig:attempt1_stem}
    \end{subfigure}%
    \begin{subfigure}[ht]{0.30\textwidth}
        \centering
        \includegraphics[height=4.45cm]{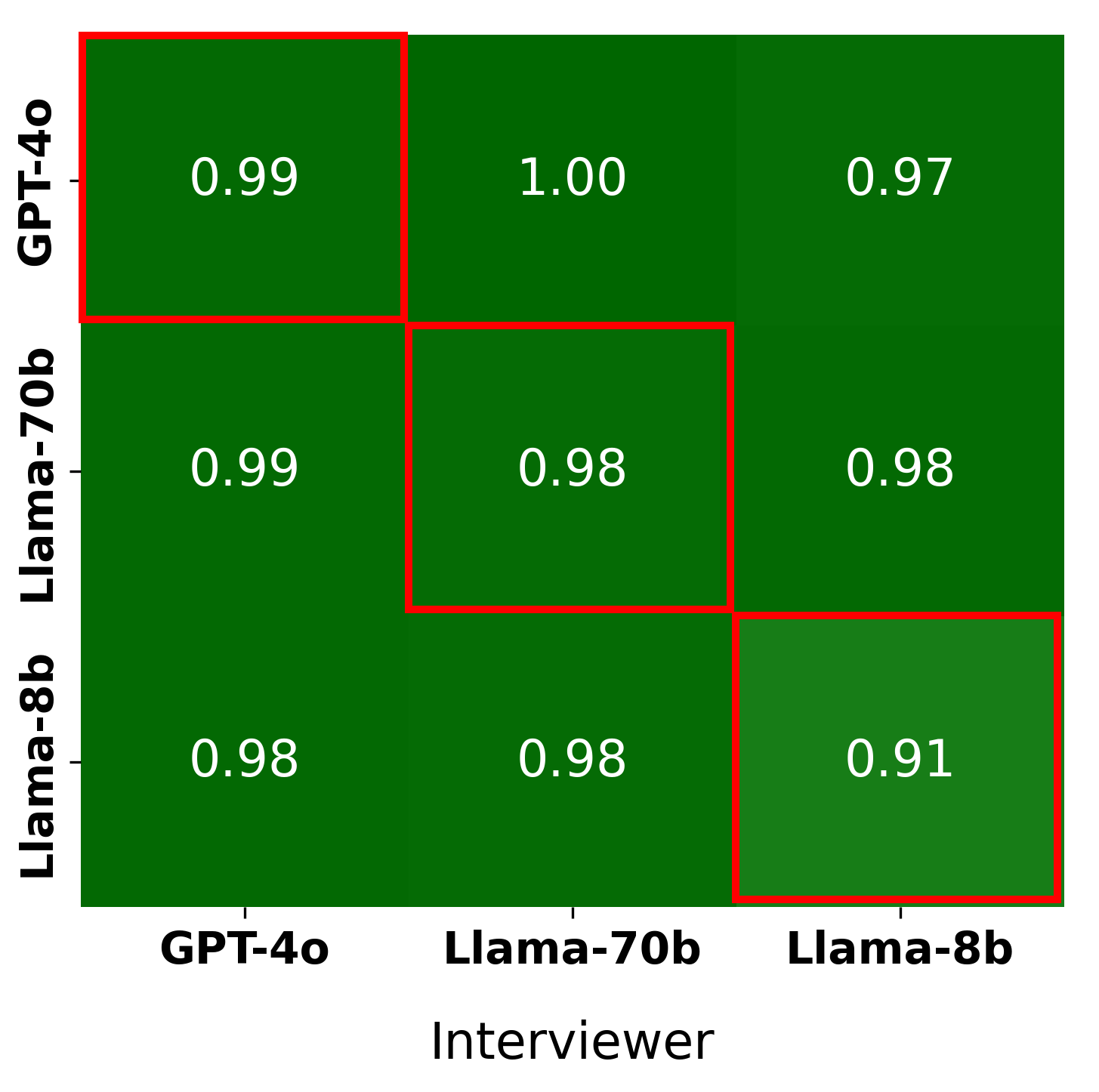}
        \caption{Interaction 2}
        \label{fig:attempt2_stem}
    \end{subfigure}%
    \begin{subfigure}[ht]{0.34\textwidth}
        \centering
        \includegraphics[height=4.45cm]{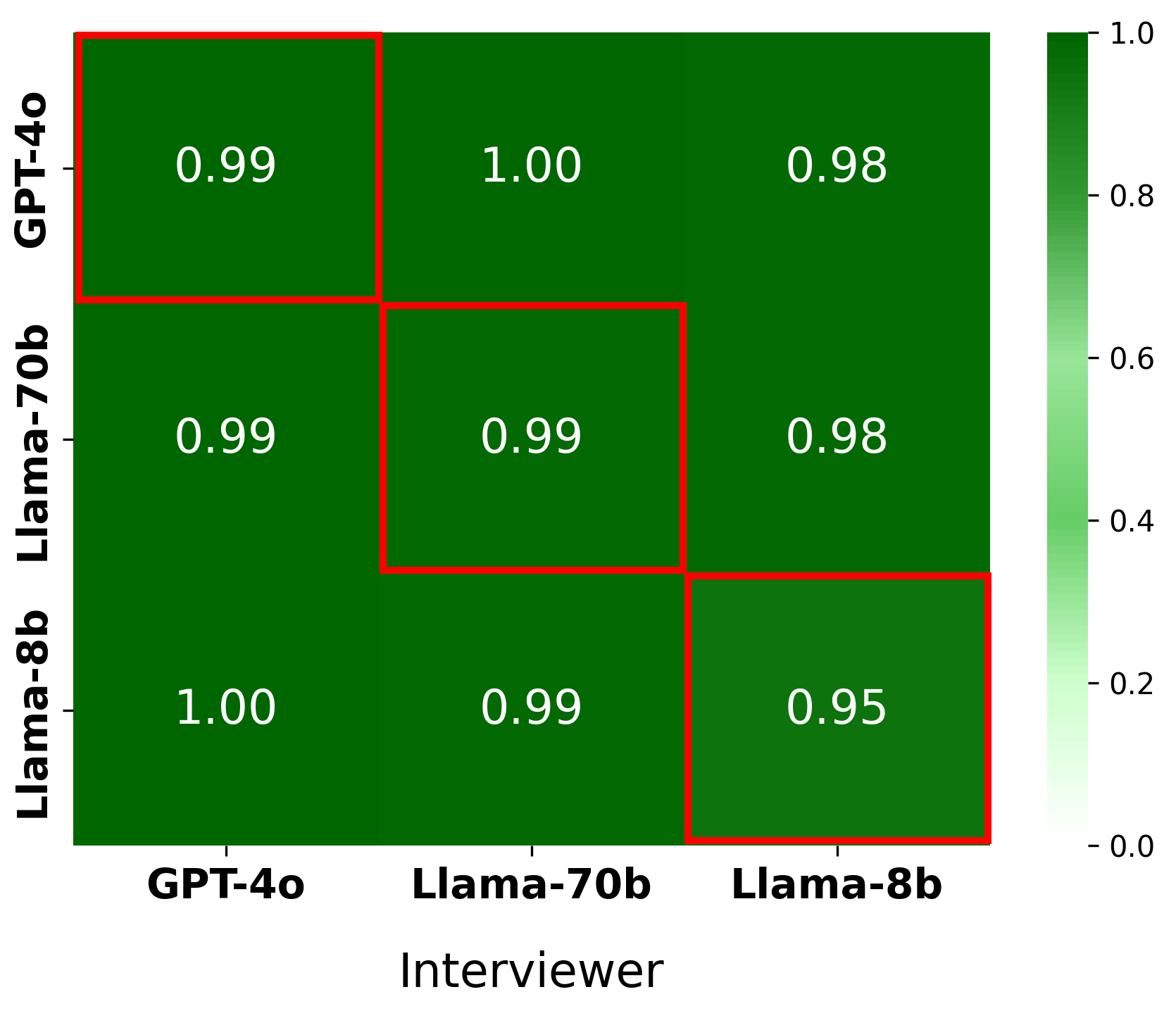}
        \caption{Interaction 3}
        \label{fig:attempt3_stem}
    \end{subfigure}
    \caption{Performance Score(Precision) by Number of Interactions in DepthQA. The red-highlighted boxes indicate self-evaluated cases. }
    \label{fig:self_bias_stem}
\end{figure*}

We analyze self-enhancement bias using GPT-4o, Llama-3.1-70B, and Llama-3.1-8B as both \Interviewer{}s and \Interviewee{}. Figures \ref{fig:self_bias}-\ref{fig:self_bias_stem} indicate that self-enhancement bias, defined as assigning higher scores to oneself, is observed only in Llama-3.1-70B among the models. In the MATH dataset, GPT-4o and Llama-3.1-8B tend to assign themselves slightly lower scores compared to other \Interviewer{}s, whereas Llama-3.1-70B assigns itself marginally higher scores. Despite this, the behavior has negligible impact on the overall rankings. As interactions progress, this bias diminishes further, and the score gap between GPT-4o and Llama-3.1-70B increases significantly. A similar pattern of diminishing bias and widening score differences is observed in the DepthQA dataset.

\begin{figure*}[ht]
    \centering
    \begin{subfigure}[ht]{0.34\textwidth}
        \centering
        \includegraphics[height=4.45cm]{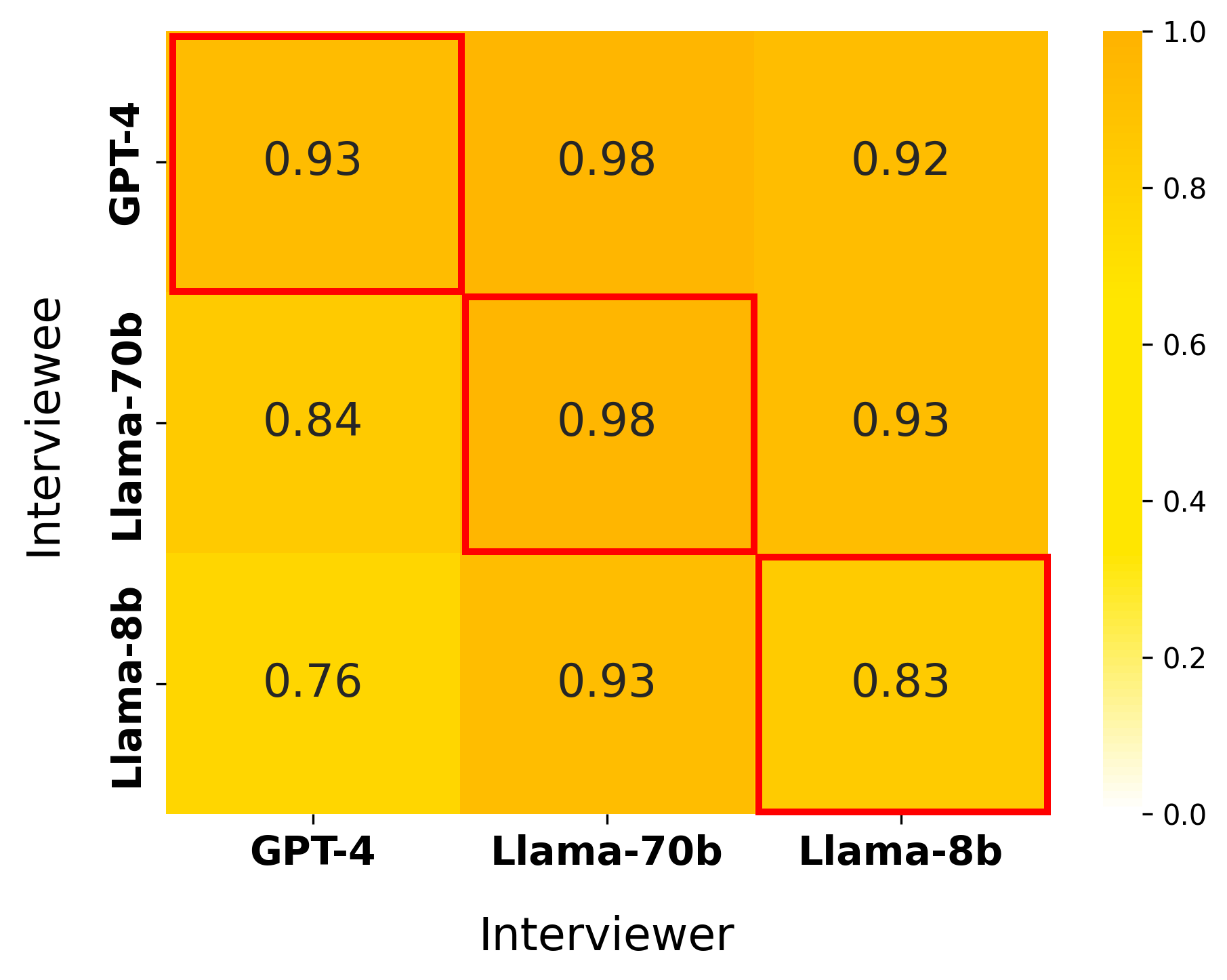}
        \caption{MATH}
        \label{fig:follow_math}
    \end{subfigure}%
    \begin{subfigure}[ht]{0.30\textwidth}
        \centering
        \includegraphics[height=4.45cm]{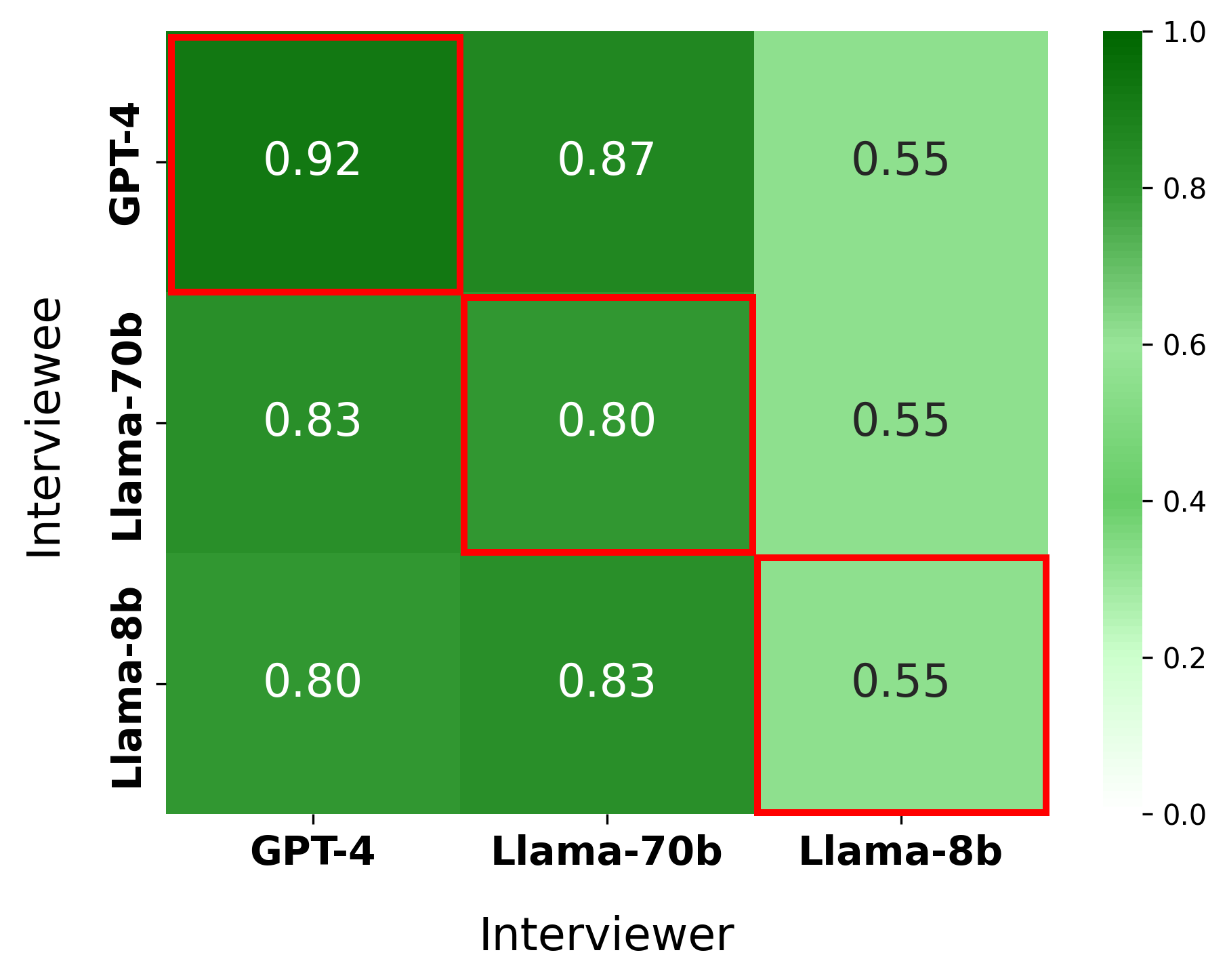}
        \caption{DepthQA}
        \label{fig:follow_stem}
    \end{subfigure}%

    \caption{Follow-Up Accuracy in MATH and DepthQA. The red-highlighted boxes indicate self-evaluated cases. }
    \label{fig:self_bias_follow}
\end{figure*}

Figure~\ref{fig:self_bias} and \ref{fig:self_bias_stem} show the trend of self-enhancement bias in MATH and DepthQA. Similar to MATH~(Figure~\ref{fig:self_bias}), no significant self-enhancement bias is observed except for Llama-3.1-70b. Figure~\ref{fig:self_bias_follow} illustrates the trends in MATH and DepthQA for follow-up questions. In MATH, as in previous results, Llama-3.1-70b consistently gives itself higher scores when grading its own responses compared to other models. However, Llama-3.1-70b also tends to give higher scores when grading other models, making it difficult to determine whether this is due to self-enhancement bias or because it generates easier follow-up questions compared to other models. In DepthQA, GPT-4 shows a similar pattern, giving itself higher scores when grading its own responses, but also assigning higher scores to other models when acting as the interviewer.

\clearpage

\end{document}